\documentclass[letterpaper, 10 pt, conference]{ieeeconf}  % Comment this line out if you need a4paper

\IEEEoverridecommandlockouts                              % This command is only needed if 
\usepackage{amsmath}
\usepackage{amssymb}
\usepackage{cite}
\usepackage{graphicx}
\usepackage{url}
\usepackage{hyperref}
\usepackage{cleveref}
\usepackage{caption}
\usepackage{subcaption}
\usepackage[nolist,printonlyused]{acronym}

\Crefname{equation}{Eq.}{Eqs.}
\Crefname{figure}{Fig.}{Figs.}
\Crefname{tabular}{Tab.}{Tabs.}
\Crefname{table}{Tab.}{Tabs.}

\title{\LARGE \bf
Towards Packaging Unit Detection for Automated Palletizing Tasks
}

\author{ \href{https://orcid.org/0009-0003-0209-8750}{Markus Völk\hspace{1mm}\includegraphics[scale=0.07]{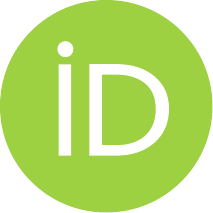}}, Kilian Kleeberger, Werner Kraus, Richard Bormann% <-this % stops a space
\thanks{The authors are with the department of Robot and Assistive Systems, Fraunhofer Institute for Manufacturing Engineering and Automation IPA, 
	Nobelstraße 12, 70569 Stuttgart
	{\tt\small markus.voelk@ipa.fraunhofer.de}}
}

\begin{document}

\maketitle
\thispagestyle{empty}
\pagestyle{empty}

\begin{abstract}
For various automated palletizing tasks, the detection of packaging units is a crucial step preceding the actual handling of the packaging units by an industrial robot. We propose an approach to this challenging problem that is fully trained on synthetically generated data and can be robustly applied to arbitrary real world packaging units without further training or setup effort. 
The proposed approach is able to handle sparse and low quality sensor data, can exploit prior knowledge if available and generalizes well to a wide range of products and application scenarios.
To demonstrate the practical use of our approach, we conduct an extensive evaluation on real-world data with a wide range of different retail products. Further, we integrated our approach in a lab demonstrator and a commercial solution will be marketed through an industrial partner.
\end{abstract}

%TODO referenzen zu anwendungen

\section{Introduction}
\begin{figure}[t]
	\includegraphics[width=\linewidth]{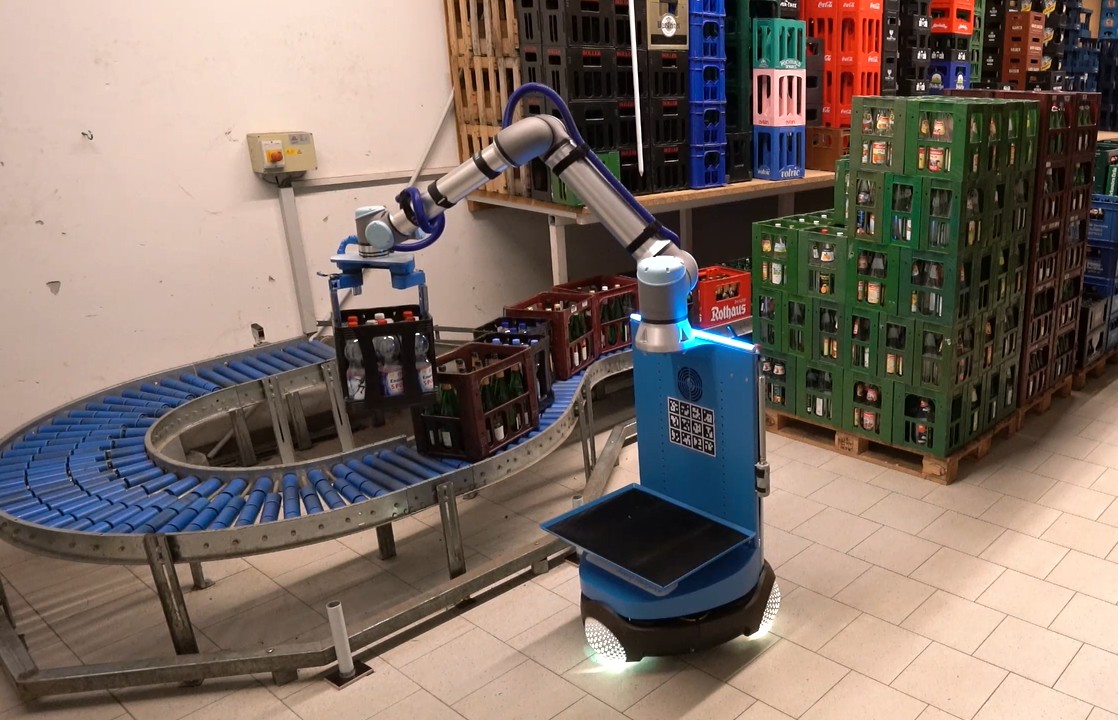}
	\caption{Beverage logistics with a mobile handling robot}
	\label{fig:luka}
\end{figure}

The task of handling packaging units appears in all kinds of logistics use cases. 
Most notable here are the distribution centers in the supply chain of retailers, supermarkets, online shops or of general postal services; practically everywhere where orders with different products have to be put together or packages need to be sorted so that they can reach their destination. Other familiar everyday applications include filling supermarket shelves or accepting empties (\Cref{fig:luka}), to name a few.
%Due to increasing labor shortages in European and American countries there is high potential for automation in all these tasks of handling packaging units.
Due to the high potential for automation in the handling of packaging units, Fraunhofer IPA has been working on this topic for several years now. It has become apparent that with a large and constantly changing product range, there are two main challenges. The first one is directly the physical manipulation of packaging units. They are usually close together and some of them, such as beverage or fruit crates, are open at the top and therefore cannot be picked with a suction or clamping gripper. As a solution to this problem, a so called roll-on-gripper\footnote{\url{https://www.youtube.com/watch?v=xB5pJRsk7xc}} \cite{muetherich_gripping_2010} was developed and is now commercially available via Premium Robotics GmbH.

The other major challenge and subject of this work is the detection (identification and 3D pose estimation) of the packing units. Normally, only RGBD sensor data and, in some use cases, the dimensions of the packaging units from stock data are available for this purpose. Classical methods of computer vision typically fail due to sensor quality, especially transparent or reflective products are an issue here, and they are not able to generalize sufficiently well to arbitrary packaging units. 
Even state-of-the-art CNN-based object detectors are often limited to 2D bounding box regression and classification. They require large amounts of annotated data and training a network for each new product is practically not feasible. In addition to the accuracy of the required annotations, the sparse depth information caused by the physical principle of most depth sensors poses another problem.

Generic \acs{CNN}-based state-of-the-art 3D object detectors \cite{hoque_comprehensive_2021} are mostly extensions of 2D single shot detectors, while some approaches are specialized for industrial applications \cite{gorschluter_survey_2022} in which RGBD sensors are available, but none of them can be directly used for general packaging unit detection.
In this paper we address these challenges and present an industrially applicable framework for the most relevant scenarios of packaging unit detection.
We further distinguish between \textit{homogeneous pallet stacks}, with only one kind of product and \textit{heterogeneous pallet stacks} which contain several different packaging units or products. We refer to the dimension (long, short, height) of the packing units as box size and consider the following scenarios. 
\begin{itemize} %TODO dictionary?
	\item Homogeneous pallet stacks with known box size: This is typically the case when orders of supermarkets etc. have to be fulfilled and heterogeneous pallet stacks have to be build in distribution centers. An additional issue in this case are intermediate layers (\textit{interlayers}) which are layers of cardboard between the packaging units to stabilize the stack. They typically have to be detected and removed before the next layer can be handled.
	\item Heterogeneous pallet stacks and box size is not known: This is the case when packaging units have to be identified for delivery services, accepting empties etc.
	\item Heterogeneous pallet stacks with known box size: Searching for boxes on a pallet stack for restocking shelves in a supermarket or retail shop for instance.
\end{itemize}
The main contributions of this paper are as follows:
\begin{itemize}

	\item An industrial grade framework/solution for the challenging task of packaging unit detection for palletizing tasks. The approach generalizes to arbitrary real world packaging units and objects of cuboid shape without any additional training.
	%We introduce a parametric generative data synthesizer for the creation of unlimited synthetic training data of all kinds of packaing units.
	\item A valuable auxiliary prediction task for single-shot object detectors.
	\item A dynamically scaling loss function with fast convergence that can be applied to classification and regression tasks as well, while it can handle the influence of outliers and simultaneously large class imbalance.
	\item A general mechanism for estimating the prediction quality of neural networks in general and especially for single-shot object detectors.
	\item An extensive evaluation of our detection framework on a depalletizing task and its application to various beverage logistics scenarios (\Cref{fig:luka}).
\end{itemize}

\iffalse
The rest of this paper is structured in the following form. In \Cref{sec:related} we give a small overview of related work on which we build in the subsequent sections. \Cref{sec:contributions} contains our contributions to generic object detectors which we later in \Cref{sec:packaging} use in our proposed packaging unit detection framework. In \Cref{sec:application} we take a close look on the application of our proposed framework to different real world scenarios and provide an extensive performance evaluation on human annotated test data. Finally we conclude \Cref{sec:conclusion} with a short discussion and figure out directions for further work.
\fi

%box size from stock data, valuable prior, prior knowledge
% related
% 3d bounding box detection, autonomious driving, kiti, etc.
% Sensor invariance, orthograpic projection

\section{Related Work}
\label{sec:related}

\subsection{Packaging Unit Detection}
\label{sec:related_packaging}
Publications on the challenging problem of general packaging unit detection are scarce. Most of them focus on depalletizing and are limited to special cases like cardboard parcel boxes \cite{monica_detection_2020,aleotti_toward_2021} or are based on model-driven bin picking \cite{holz_real_time_2015}. The typical approach to this problem is based on classical \ac{CV} for feature detection \cite{arpenti_rgbd_2020,li_workpiece_2020} and solving a combinatorial optimization problem \cite{arpenti_rgbd_2020,monica_detection_2020}. Others build on even stronger simplifications such as RFID-Tags \cite{prasse_concept_2011} or focus on hardware and the overall architecture \cite{caccavale_flexible_2020}.
Non of these approaches is able to generalize to arbitrary packing units in homogeneous and heterogeneous pallet stacks while being robust against sensor limitations and environmental conditions. Our proposed framework for packaging unit detection fulfills all this requirements. It generalizes to arbitrary packing units or objects of cuboid shape without the need of further training or manual setup and can perform the detection in less then 100~ms.

\subsection{Relevant Learning Techniques}
\label{sec:related_learning}
Single-shot object detectors like \acs{SSD} \cite{liu_ssd_2016}, \acs{YOLO} \cite{bochkovskiy_yolov4_2020, wang_yolov7_2022} or RetinaNet \cite{lin_focal_2017} are current state-of-the-art in terms of prediction quality and speed.
They are typically fully conventional \acp{CNN} that make local predictions for object classification and a bounding box regression, which are subsequently filtered in post-processing step called \ac{NMS}. Their training with basic loss functions such as the \ac{BCE} and L2 loss face three main issues. The first issue is the imbalance of classes in the local classification ground truth. This can either be addressed by a sampling strategy known as \ac{OHEM} \cite{shrivastava_training_2016} or more recently by loss functions like the \ac{FL} \cite{lin_focal_2017}, the \ac{RFL} \cite{sergievskiy_reduced_2019} or the \ac{SL} \cite{lu_deep_2018} which automatically weight down easy samples based on their absolute error. 

The second issue are outliers resulting from low quality bounding box annotations done by humans. One approach to reduce their influence is a combination between L1 and L2 loss called the Huber Loss \cite{huber_robust_1964} or more popular its special case the Smooth L1 Loss \cite{girshick_fast_2015}. The third issue also considers bounding box regression and arises from the large variance in object scale. Recent approaches like the Distance IoU Loss \cite{zheng_distance-iou_2019} solve this issue by utilizing scale independent metrics like the \ac{IoU}. Other recent approaches \cite{tian_fcos_2019,wu_iou-aware_2020} perform quality estimates which are later used in the post-processing step to suppress bounding boxes with low quality. Two of these quality estimates are the center-ness score \cite{tian_fcos_2019} and the \ac{IoU} score \cite{wu_iou-aware_2020}, whereby the predicted \ac{IoU} is an estimate derived from an other estimate, the bounding box prediction. In \Cref{sec:contributions} we propose a unified loss function that addresses all these issues and a generalized quality estimate for arbitrary local predictions.

The last years have brought some improvements and specializations concerning the convolution operation and the architecture of \acsp{CNN}, which we will use in \Cref{sec:packaging}. Separable Convolutions \cite{chollet_xception_2016, howard_mobilenets_2017} are used in Inverted Residual blocks or MBConv blocks \cite{sandler_mobilenetv2_2018} to heavily (90\% or more) reduce parameter count without notable performance loss. Adding the spatial location on which the convolution kernel is applied on the feature map as additional features to convolution input \cite{liu_intriguing_2018} improves the performance for practically all tasks which require more spatial or contextual information. Sparse \cite{uhrig_sparsity_2017} and Partial Convolution \cite{liu_image_2018} where proposed to handle sparse input directly in the network by propagating the sparsity signal through the network and considering it in the convolution operation.
Furthermore, it has been shown \cite{hinterstoisser_annotation_2019} that object detectors trained on synthetically generated data can outperform ones trained on real images under the right circumstances and that \ac{PBR} \cite{hodan_photorealistic_2019} outperforms other methods of synthetic data generation.
Also, adding additional sparsity to the input data is not necessarily a harmful undertaking. It can be used as valuable auxiliary task for learning  representation for language \cite{devlin_bert_2019} or vision \cite{he_masked_2021} task and works as regularization like in the well known Dropout \cite{srivastava_dropout_2014}.

\section{General Contributions}
\label{sec:contributions}
This section introduces some novel concepts for generic single-shot detectors which we will later use in our proposed detection framework.

\subsection{Bounded Distance Transform}
\label{sec:tranform}
\begin{figure}[tb]
	\includegraphics[width=\linewidth]{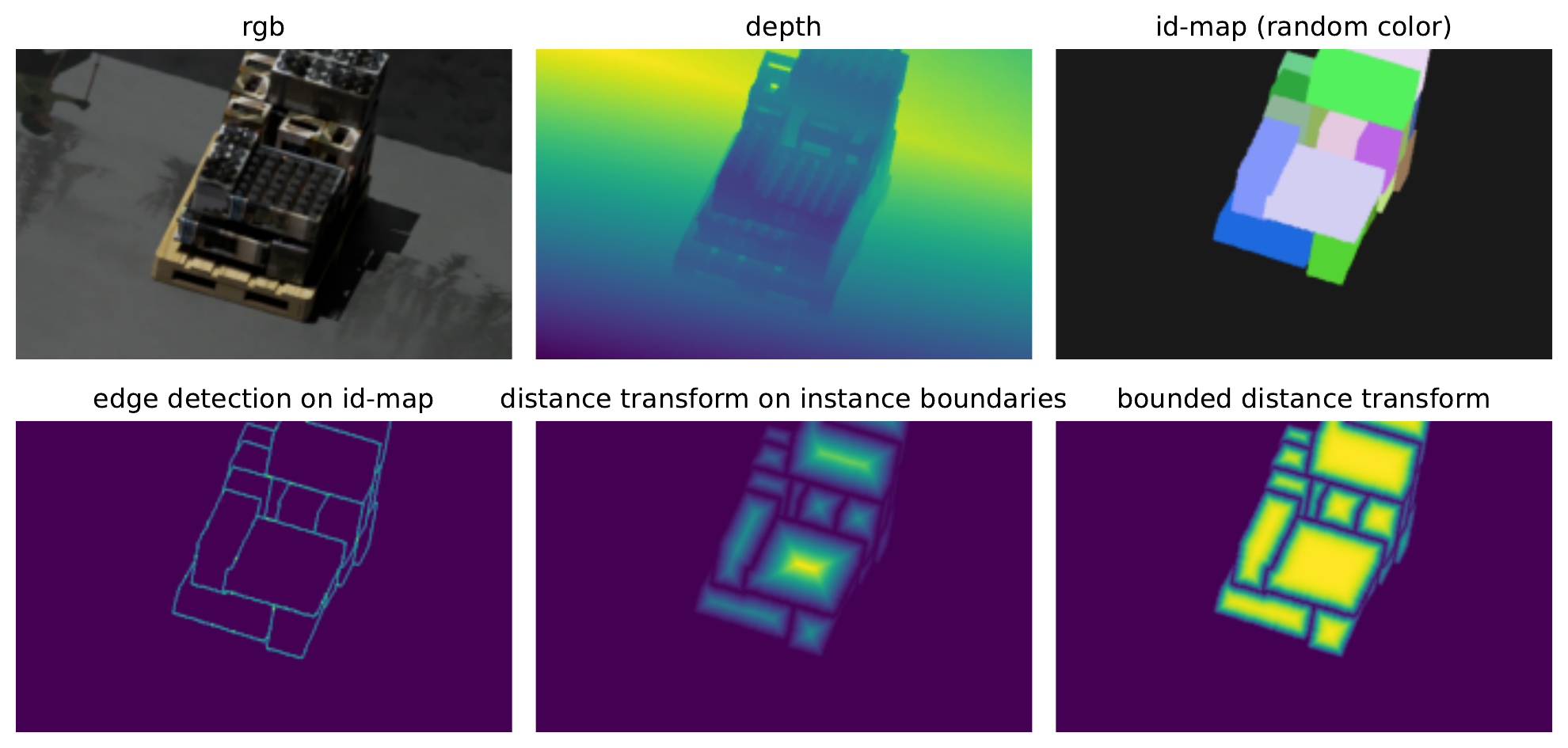}
	\caption{\acl{BDT} (rendered image data with orthographic camera)}
	\label{fig:distance_transform}
\end{figure}
We propose an auxiliary prediction task that can be used in the post-processing step (\ac{NMS}) to efficiently separate instances and suppress uncertain predictions near the instance boundary.
In the following, we call the ground truth for this task and its computation the \ac{BDT}. The steps for obtaining the \ac{BDT} $d_b$ from id-maps or segmentation ground truth are visualized in \Cref{fig:distance_transform}. First we use Sobel filters for detecting the instance boundaries from the id-maps. Second we perform the Euclidean Distance Transform \cite{strutz_distance_2021} on the edges representing the instance boundaries and set the background areas to zero. As third and last step we use the $\tanh$ to bring the values into the interval $[0,1]$ and keep them near $1$ on the inside of the instances.
\begin{equation}
	\label{eq:dist}
	d_b = \tanh{\frac{d}{s}}
\end{equation}
Here, $d$ refers to the Euclidean Distance Transform and $s$ to a scaling factor which we choose based on the down-sampling ratio of our \ac{CNN}-architecture to get suitable ground truth.

Predictions of the \ac{BDT} can later in the post-processing be used for thresholding while the $\tanh$ prevents small instances from being ignored. It should also be possible to directly use a prediction of the \ac{BDT} for instance segmentation. For this it would make sense to perform a morphological operation with suitable kernel size to compensate the too small instance masks resulting from thresholding.
All the necessary operations for the \ac{BDT} can be done using classical \acs{CV} libraries like OpenCV \cite{noauthor_opencv_2014} and to the best of our knowledge, this simple but efficient idea was not published before.

\subsection{Dynamically Scaled Shrinkage Loss}
\label{sec:loss}
In this section we propose a loss function that dynamically scales the absolute error before passing it through its nonlinearity. Since we build on the idea of the well known \acl{SL} \cite{lu_deep_2018}, we call our proposed loss function the \ac{DSSL}. The \ac{DSSL} can be applied to classification and regression problems as well.

\begin{figure}[tb]
	\includegraphics[width=\linewidth]{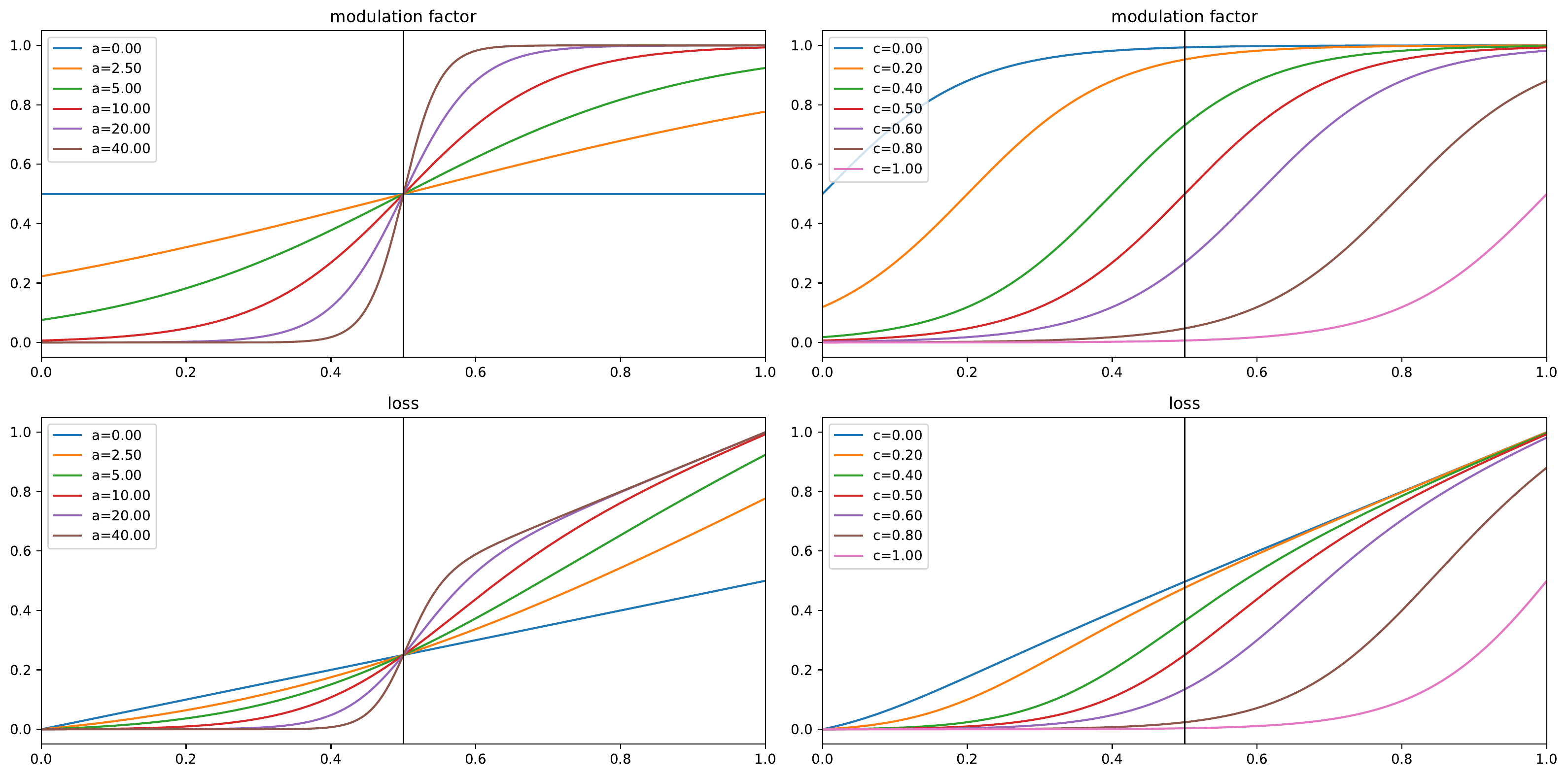}
	\caption{Influence of the parameters $a$ (left) and $c$ (right) on the modulation factor \Cref{eq:factor} (top) and the loss function \Cref{eq:dynamic} (bottom) depending on the absolute or respectively the normalized absolute error}
	\label{fig:shrinkage}
\end{figure}

\begin{figure}[tb]
	\includegraphics[width=\linewidth]{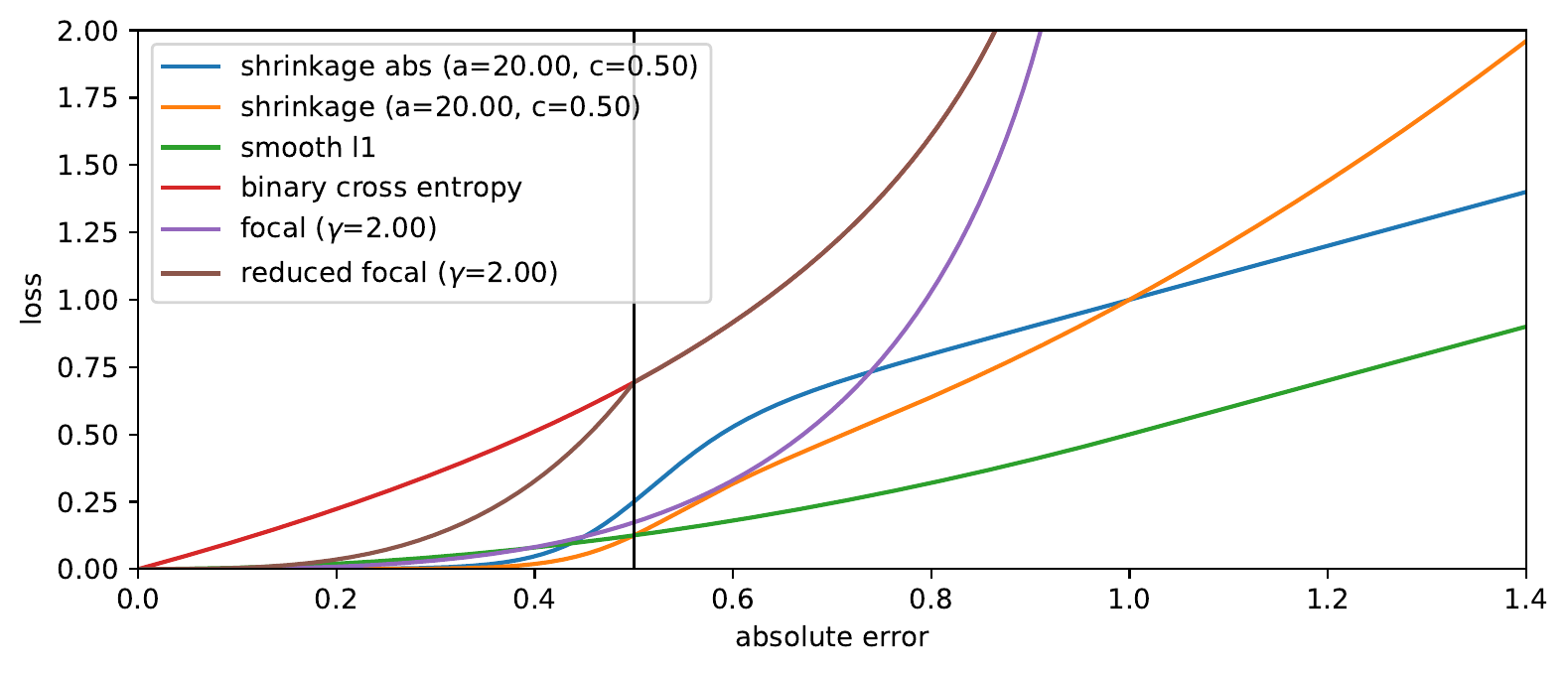}
	\caption{Comparison of various loss functions; we use 'shrinkage abs' for the \ac{DSSL}, where the mean absolute error is dynamically scaled to $0.5$}
	\label{fig:loss_comparison}
\end{figure}

In the following, we consider the local predictions on a feature map with shape $B \times H \times W \times K$, with $b$, $h$, $w$ and $k$ as the indices on this feature map. $B$ represents the batch size, $H$ and $W$ the spatial dimensions height and width and $K$ the number of considered channels, for instance, $K$ confidences for the classification of $K$ classes.
The absolute error for one element can then be written as
\begin{equation}
	\label{eq:abs}
	l_{bhwk} = \left|y_{bhwk}-\hat{y}_{bhwk}\right| 
	%l = \left|y-\hat{y}\right|
\end{equation}
where $y$ denotes the local ground truth and $\hat{y}$ the prediction of the network.
Starting with the average absolute error within a batch $\bar{l}$
\begin{equation}
	\label{eq:mean}
	\bar{l} = \frac{1}{BHWK} \sum_{b=1}^B \sum_{h=1}^H \sum_{w=1}^W \sum_{k=1}^K l_{bhwk}
\end{equation}
we can write a normalized form $n_{bhwk}$ of the absolute error
\begin{equation}
	\label{eq:norm}
	n_{bhwk} = \frac{l_{bhwk}}{2\mathrm{gs}\left(\bar{l}\right)+\epsilon}
\end{equation}
Where $\mathrm{gs}(\cdot)$ stops the gradient propagation during training. $\epsilon$ is a small constant for numerical stability.
Utilizing the modulation factor $f(l)$ with its parameters $a$ and $c$ from \acl{SL}
\begin{equation}
	\label{eq:factor}
	f(l) = \frac{1}{1+\exp{a(c-l)}}
\end{equation}
we can formulate the following local loss function
\begin{equation}
	\label{eq:dynamic}
	L_{bhwk} = \frac{n_{bhwk}}{1+\exp{a(c-n_{bhwk})}}
\end{equation}
which leads to our global loss function
\begin{equation}
	\label{eq:loss_mean}
	L_{DSSL} = \frac{1}{BHW} \sum_{b=1}^B \sum_{h=1}^H \sum_{w=1}^W \sum_{k=1}^K L_{bhwk}
\end{equation}
%variant of shrinkage loss, $l_{bhwc}$ in the numerator, not $l_{bhwc}^2$
%\begin{equation}
%\label{eq:shrinkage}
%L_{bhwc} = \frac{l_{bhwc}}{1+\exp{a(c-l_{bhwc})}}
%\end{equation}
The proposed \ac{DSSL} has some notable properties. The dynamic scaling of the absolute errors has the consequence that the loss function always operates around its inflection point at $0.5$. Replacing $n_{bhwk}$ by $l_{bhwk}$ in \Cref{eq:dynamic} disables the dynamic scaling behavior. Instead of the squared absolute error in the \ac{SL}, we use the absolute respectively the normalized absolute error in the nominator of \Cref{eq:dynamic}. This avoids a second nonlinearity, reduces the influence of outliers and makes things more interpretable. \Cref{fig:shrinkage} shows the influence of the parameters $c$ and $a$ on the modulation factor and the loss function itself\footnote{These plots may also be useful for choosing the parameters in practice.}.
\Cref{fig:loss_comparison} compares \ac{DSSL} with other loss functions. It has the property of down-weighting the easy samples as it is done by the \ac{FL} while at the same time reducing the influence of outliers as it is done by the Smooth L1 Loss. Smooth L1 and \ac{FL} have also the drawback of nearly vanishing gradients for small regression errors. Due to the dynamic scaling we always get strong error signals and let the training dynamics do the rest while we reduce the learning rate during training.
Another consequence of the normalization is that the loss value no longer decreases during training (\Cref{fig:log_history}). Therefore, the absolute error may be a better choice for monitoring the training progress.
One may also consider to maintain a moving average of the mean absolute error, but in our experiments, we found that the statistic over batch and spatial dimensions are sufficient for fast and stable convergence.
If we apply the \ac{DSSL} with our default parameters ($a=20$ and $c=0.5$) to a classification task like the one shown in \Cref{fig:prediction_sim,fig:prediction_real} (background, short, long), it rapidly fits the binary labels. 
For smoother values that look more like a probability distribution as produced by the \ac{FL}, it would make sense to choose the parameters $a$ and $c$ more suitable, but for most classification tasks simply the $\mathrm{argmax}$ function is used. Also for bounding box regression it may be desired to get an error signal that is independent of the actual object scale, as with \ac{IoU} Loss (\Cref{sec:related_learning}). This can be achieved by dividing the absolute error after \Cref{eq:abs} by the object scale.
We provide our implementation\footnote{\url{https://github.com/mvoelk/ssd_detectors/blob/master/utils/losses.py}} of \ac{DSSL} in TensorFlow.

%after training, we would have absolute values << 0.5 and they all will be weighted down and vanishing gradients

\subsection{Prediction of Certainty}
\label{sec:certainty}
\begin{figure}[tb]
	\includegraphics[width=\linewidth]{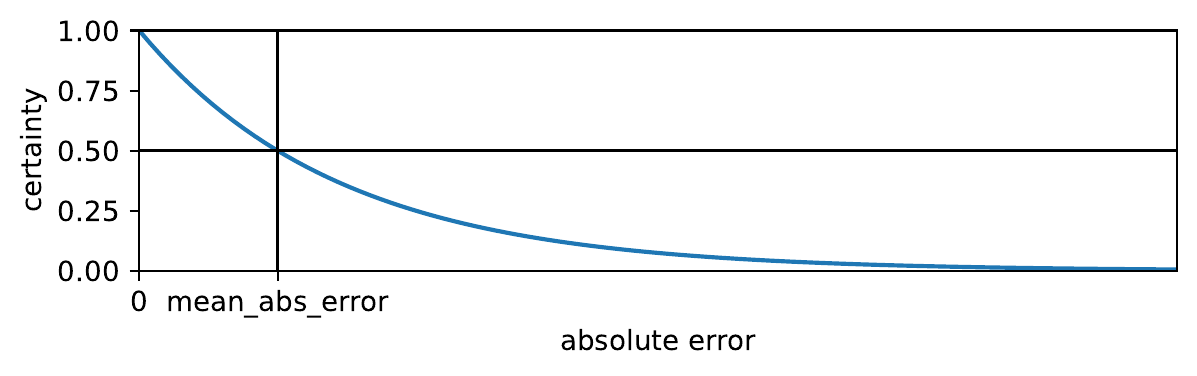}
	\caption{Certainty function as defined in \Cref{eq:certainty}}
	\label{fig:ceratainty}
\end{figure}

\begin{figure}[bt]
	\centering
	\includegraphics[width=\linewidth]{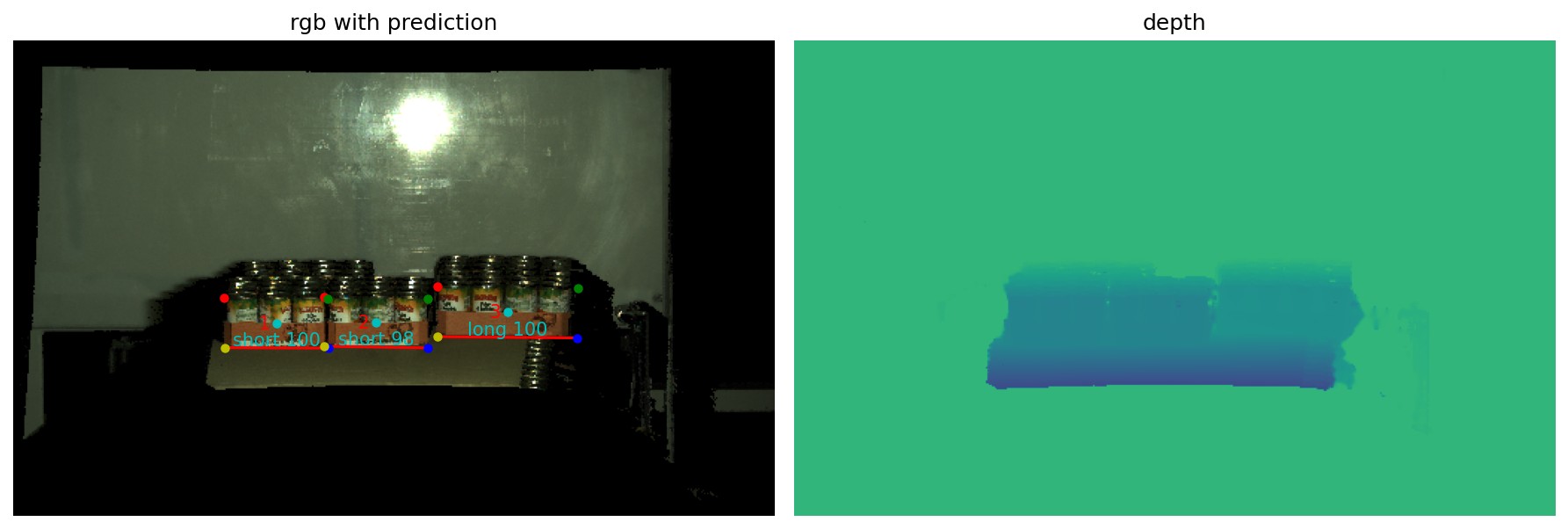}
	\includegraphics[width=\linewidth]{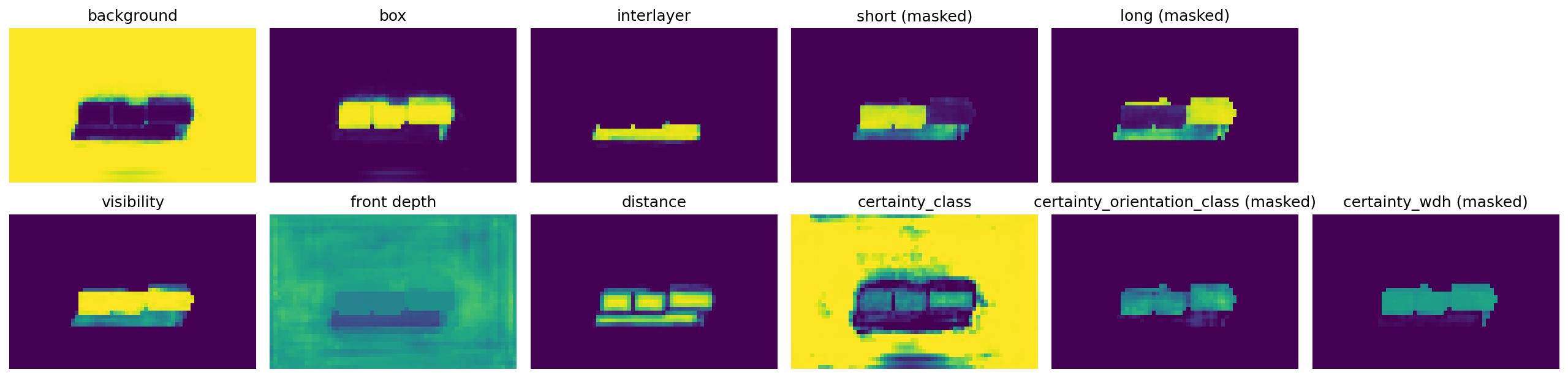}
	\caption{Prediction on orthographic projected real data}
	\label{fig:prediction_real_orthograpic}
	\vspace{-4mm}
\end{figure}

In this section we propose a generalization of the quality estimate mentioned in \Cref{sec:related_learning}. The basic idea here is simple, let the network estimate its own prediction error.
A direct prediction of the regression error as a quantitative measure may be difficult to handle in a post-processing (\ac{NMS}) step, at least if a certain threshold has to be determined. Therefore, based on \Cref{eq:abs} and \Cref{eq:mean}, we propose a unified quality measure \Cref{eq:certainty} which is defined in the interval $[0,1]$ and where higher is better.
\begin{equation}
	\text{certainty}_{bhwk} = \exp\left( \frac{l_{bhwk}}{\mathrm{gs}\left(\bar{l}\right)+\epsilon } \log\frac{1}{2}\right)
	\label{eq:certainty}
\end{equation}
The plot in \Cref{fig:ceratainty} shows a graph of the proposed certainty function, which is $0.5$ for the average error, $1$ for an error of $0$ and $0$ for an infinitely large error.
As our intuition suggests and \cite{wu_iou-aware_2020} have shown, the backpropagation of the gradient resulting from the actual prediction is not beneficial for the overall performance. Therefore we also stop the gradient propagation at this place.
The training can be done via the \ac{DSSL} proposed in \Cref{sec:loss}.
%In some situations (position prediction with $x,y,z$ values for instance) it may makes sense to perform certain operations like an L2 norm before calculating the certainty. In other situations it may make more sense to estimate the certainty over all $K$ channels and take the minimum.
Our first experiments in \Cref{fig:prediction_real_orthograpic} indeed show that the predicted certainty is low in areas with insufficient information or ambiguity due to partially hidden objects etc. %We weighted down the loss terms for the certainty prediction with a value of $0.1$. This was more or less done by intuition, since the certainty is some kind of second order prediction and at the time of writing, we do not know if this is beneficial or even necessary.

% TODO relation probability vs certainty

\section{Packaging Unit Detection}
\label{sec:packaging}

\subsection{Data Generation}
\label{sec:genrator}

Object detectors $\hat{f}(x)$ in general, approximate a function $y = f(x)$ based on a finite number of training samples $(X_n,Y_n)_{n=1}^{N}$, where $y$ denotes the abstract representation of the objects (category, position, orientation, ...) and $x$ denotes the sensor data corresponding to the scene. Since human annotations on real 3D sensor data are laborious and often inaccurate, it is much easier and more flexible to model the inverse $f^{-1}(y)$ of this function using computer graphics and sample training data from this simulated distribution.

\begin{figure}[tb]
	\includegraphics[width=\linewidth]{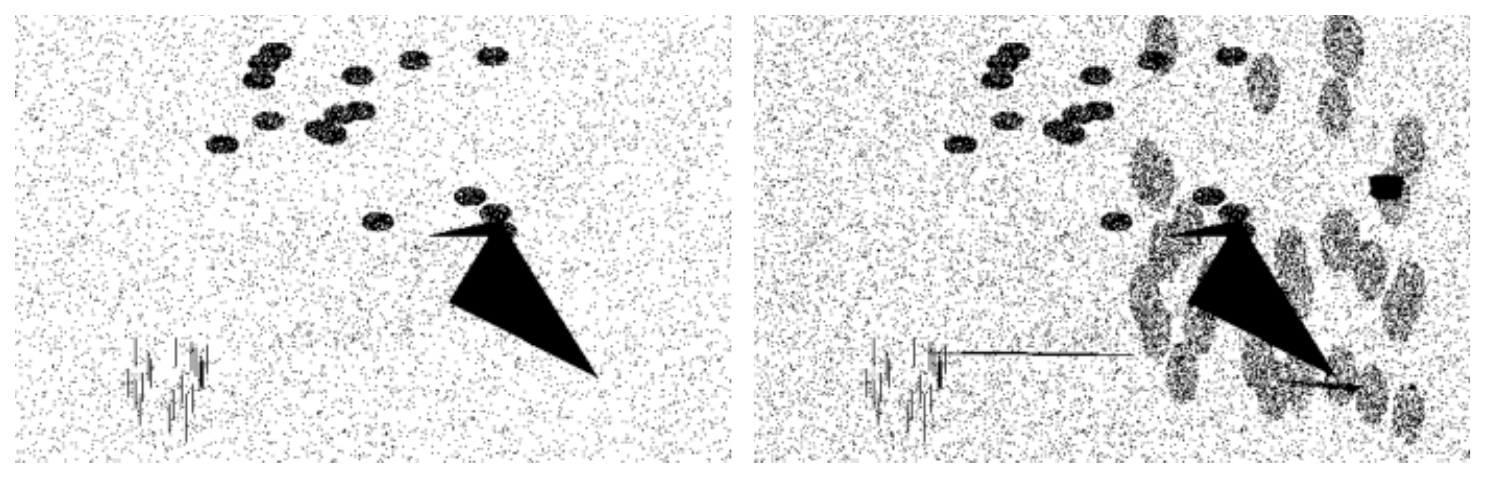}
	\caption{Random artificial sparsity for RGB image (left) and with additional sparsity on depth image (right)}
	\label{fig:sparsity}
\end{figure}

\begin{figure}[tb]
	\centering
	\begin{subfigure}[b]{\linewidth}
		\includegraphics[width=\linewidth,trim={0 0 0 3cm},clip]{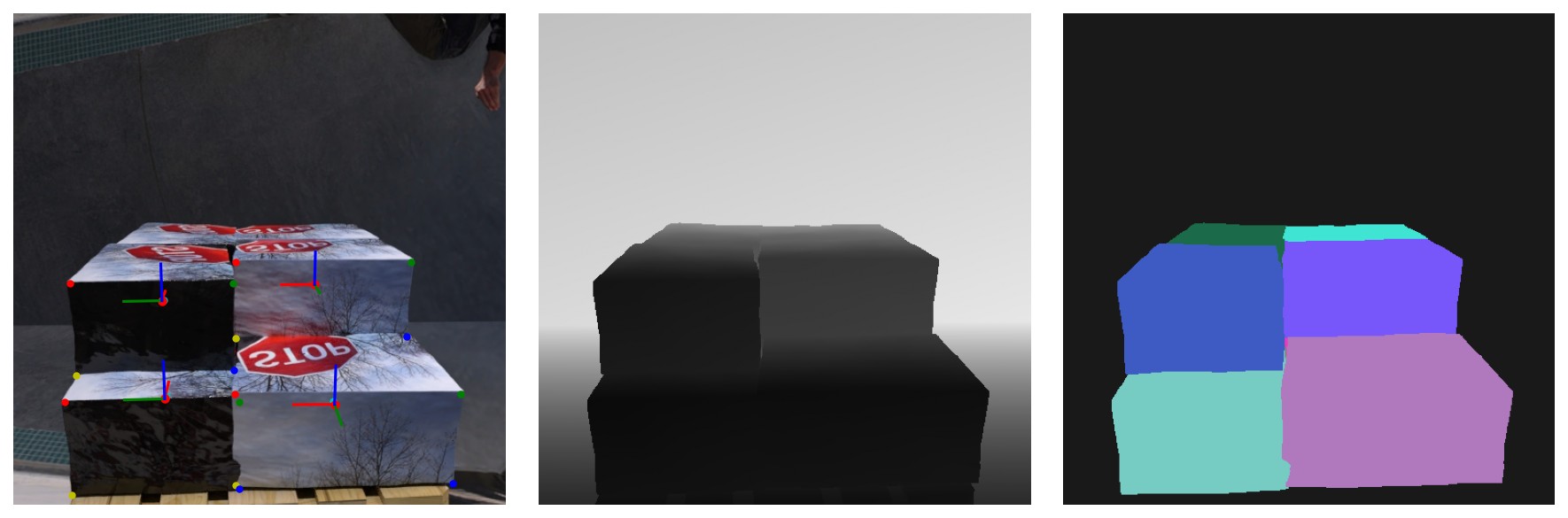}
	\end{subfigure}
	\vfill
	\begin{subfigure}[b]{\linewidth}
		\includegraphics[width=\linewidth,trim={0 0 0 3cm},clip]{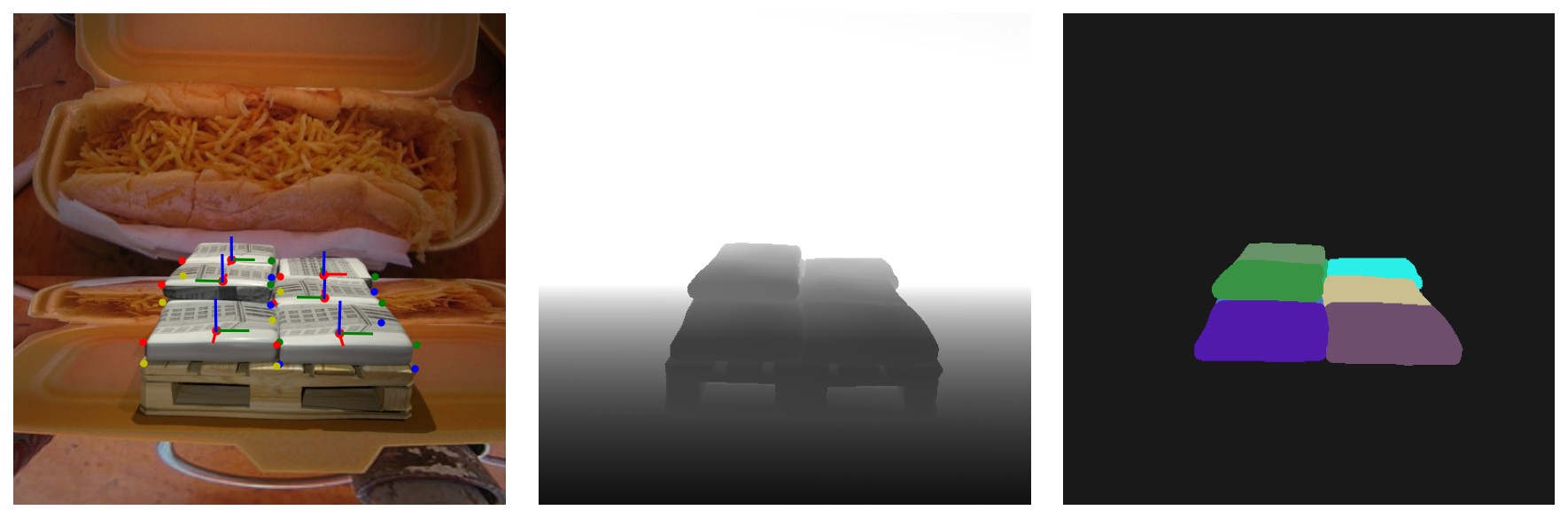}
	\end{subfigure}
	\vfill
	\begin{subfigure}[b]{\linewidth}
		\includegraphics[width=\linewidth]{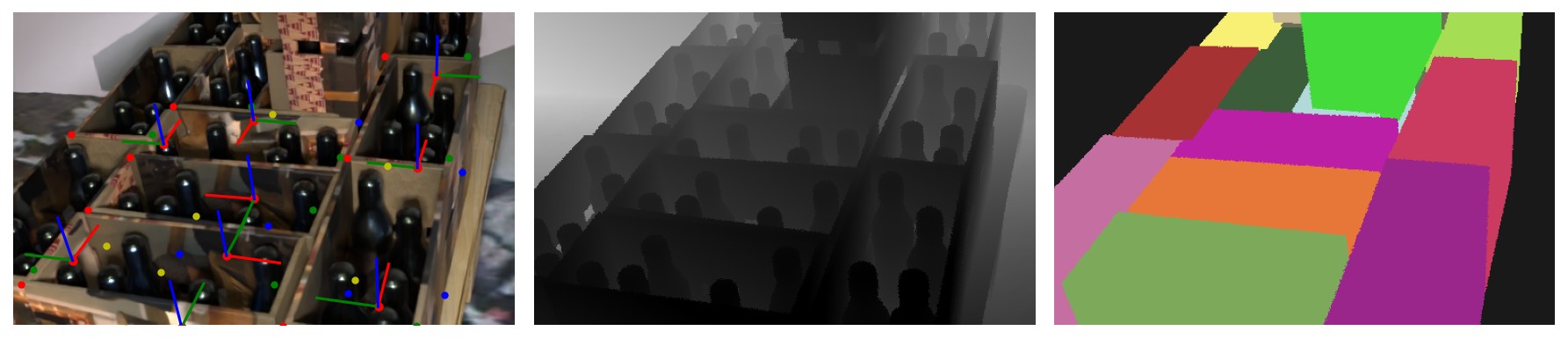}
	\end{subfigure}
	\vfill
	\begin{subfigure}[b]{\linewidth}
		\includegraphics[width=\linewidth]{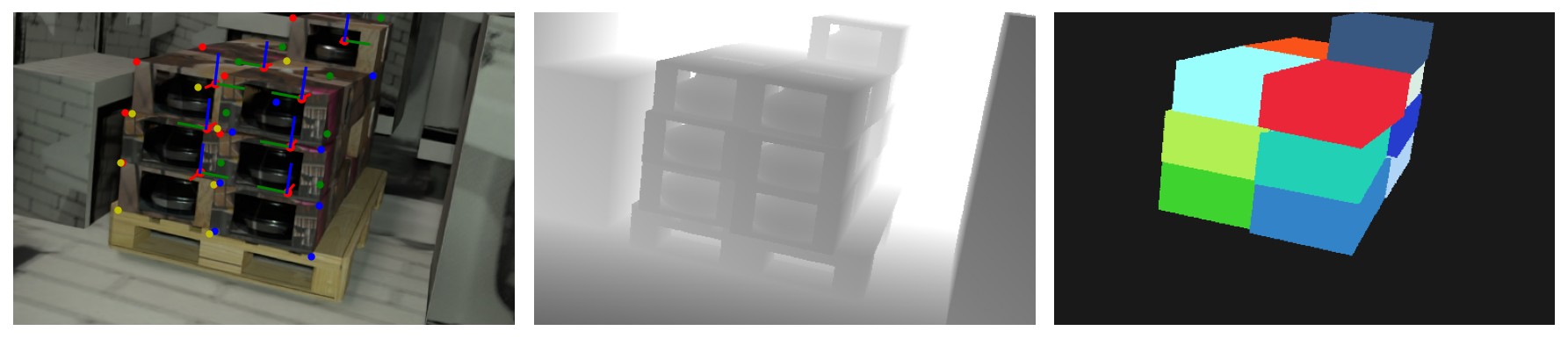}
	\end{subfigure}
	\caption{Synthetically generated data; RGB image with (left), depth image (middle) and segmentation ground truth / id-map (right), detection ground truth (box frame and keypoints) drawn in RGB image}
	\label{fig:ground_truth}
\end{figure}

We utilize Blender\footnote{\url{https://www.blender.org}} to render photorealistic scenes of pallet stacks and get the corresponding ground truth (\Cref{fig:ground_truth}) without manual annotations.
During this process of procedural data generation we randomize certain scene properties like camera view, lighting, background as well as the packaging pattern and the size and type of the packaging units themselves. We model certain specific types of packaging units like, boxes, bags, beverage crates and closed and open cartboard boxes (fruit and vegetables) etc., as well as their content in form of cans, bottles, packages (beverage cartons, cardboard packages, etc.) or random stuff. 
For the packaging units, their content, the background and additional disturbance geometry, we excessively randomize the geometry using tessellation, the texture and the material properties in form of the shader parameterization.
To achieve robustness against the noise and sparsity resulting from sensor limitations we add several artifacts to the synthesized sensor data.
We add Gaussian noise with random standard deviation to mimic sensor noise and Gaussian blur with random kernel size and standard deviation to mimic blur resulting from various intrinsic filtering techniques. In addition we add artificial sparsity as shown in \Cref{fig:sparsity}. These binary masks are a combination of binary noise drawn from a Bernoulli distribution, dilated binary noise and random polygons and elliptic shapes that mimic large sharp and round sparse areas resulting from the physical measuring principle of the sensors.
In summary we get the simulated sensor data in form of RGB and depth images and the ground truth for packaging unit detection in form of size, position, orientation, keypoints (top-lef, top-right, bottom-right, bottom-left corner of the front and back side of the packaging unit) in image and 3D space, visibility values and id-maps for instance segmentation. The visibility is the percentage of a packaging unit instance that is not occluded by other instances. An estimate of the visibility may later be helpful to decide whether a packaging unit can be picked or not.

% homogeneous or heterogeneous pallet stacks

\subsection{Architecture}
\label{sec:architecture}
\begin{figure}[tb]
	\includegraphics[width=\linewidth]{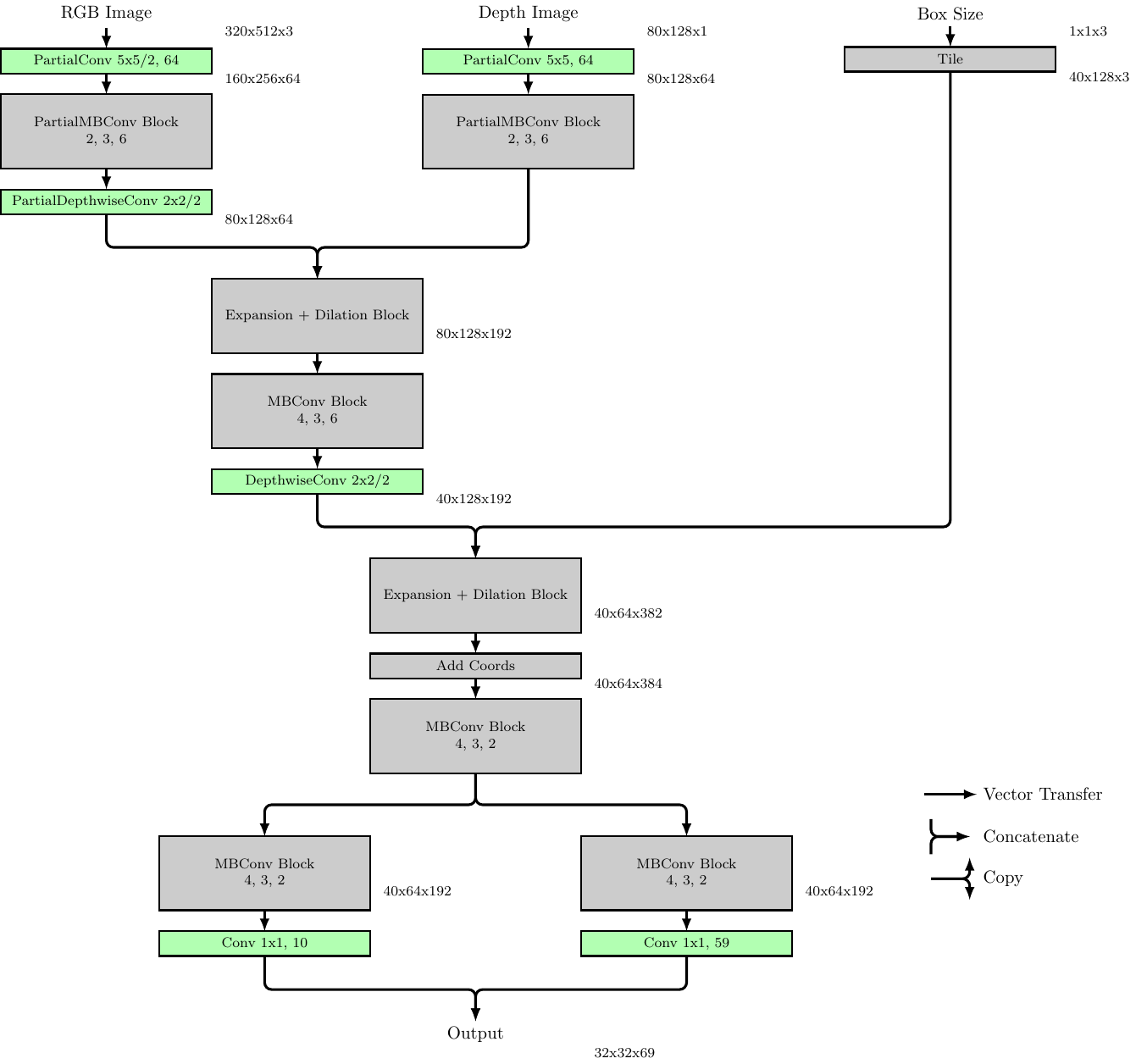}
	\caption{Architecture of the proposed network; notation for Conv: kernel\_size/stride, features; notation for MBConv: repeats, kernel\_size, expansion}
	\label{fig:architecture}
\end{figure}

For the special case of packaging unit detection, we propose a suitable fully convolutional network architecture, shown in \Cref{fig:architecture}, that directly results from the prior work mentioned in \Cref{sec:related_learning}.
We use two separate encoder based MBConv blocks combined with Partial Convolution to handle sparse input data and concatenate the features together. We then use 1x1 convolution and dilated convolution to change the number of features and aggregate more spatial information. After a second stage of MBConv blocks we concatenate the tiled box size to the feature map. We also add the spatial location of elements on the feature map as additional features \cite{liu_intriguing_2018} and end up with a third and fourth stage in two separate branches, one for the classification and one for the regression part. We use \ac{ReLU} activation functions for the whole architecture except for final output activation which we choose appropriately to the prediction (softmax, sigmod or linear).
The local predictions of the network contain a classification (background, box and interlayer), a second classification for the box orientation (short, long), a confidence for the prior (whether the detected instance fits to the prior / provided box size or not), visibility, relative key points in image space, keypoints in 3D, position and orientation (elements of a rotation matrix) in 3D, distance of the front face, box dimension (short, long, height), box dimension (width, depth, height), the \ac{BDT} proposed in \Cref{sec:tranform} and the certainty (\Cref{sec:certainty}) for classification, orientation classification and box size regression.

\subsection{Training}
\label{sec:train}
For all predictions, we use the \ac{DSSL} proposed in \Cref{sec:loss}. To handle the symmetries of packaging units along the z-axis (bottom to top) for the prediction of the orientation, we calculate the loss values for all valid orientations and select the smallest value for backpropagation.
Depending on the use case, we decide whether we use training data with homogeneous or heterogeneous pallet stacks and whether we can omit the rightmost branch in \Cref{fig:architecture} providing the box size.
It is clear, that the model will learn to make the best predictions for the case of homogeneous pallet stacks and known box size.
For the more challenging use case with heterogeneous pallet stacks and unknown box size, the model has to learn to predict the box dimensions. It may also be easier to learn the box dimensions in form (width, depth, height) in camera view instead of (short, long, height), where rotation comes into play and the box prediction may totally fail due to a wrongly estimated depth dimension for instance. In the case of heterogeneous pallet stacks and known box size, the model will learn to distinguish the packaging units from others via the prior confidence and will also do a better job on estimating their position and orientation.

% multi taks, addional error signal

\subsection{Prediction and Post-Processing}
\label{sec:postprocessing}
In this section we consider the post processing step for filtering the dense predictions from the \acs{CNN} output. In addition to the direct pose regression for the instances, we also reconstruct their pose from the 3D keypoint regression. To select good candidates from the local predictions, we filter them based on classification, visibility, \ac{BDT} and certainty. We group the remaining candidates based on their minimum box dimension and their Euclidean distance to each other before we select the candidate that is nearest to the spatial mean of each group. Finally, we sort the remaining predictions in the pallet coordinate system which is either known directly or a rough estimate is available. We sort them suitable for the application and based on their height, from top to bottom and from near to far.
%TODO details, we filter them based on classification, ...

\section{Application}
\label{sec:application}

\subsubsection{Implementation}
\label{sec:implementation}

\begin{figure}
	\centering
	\begin{subfigure}{.5\linewidth}
		\centering
		\includegraphics[width=\linewidth]{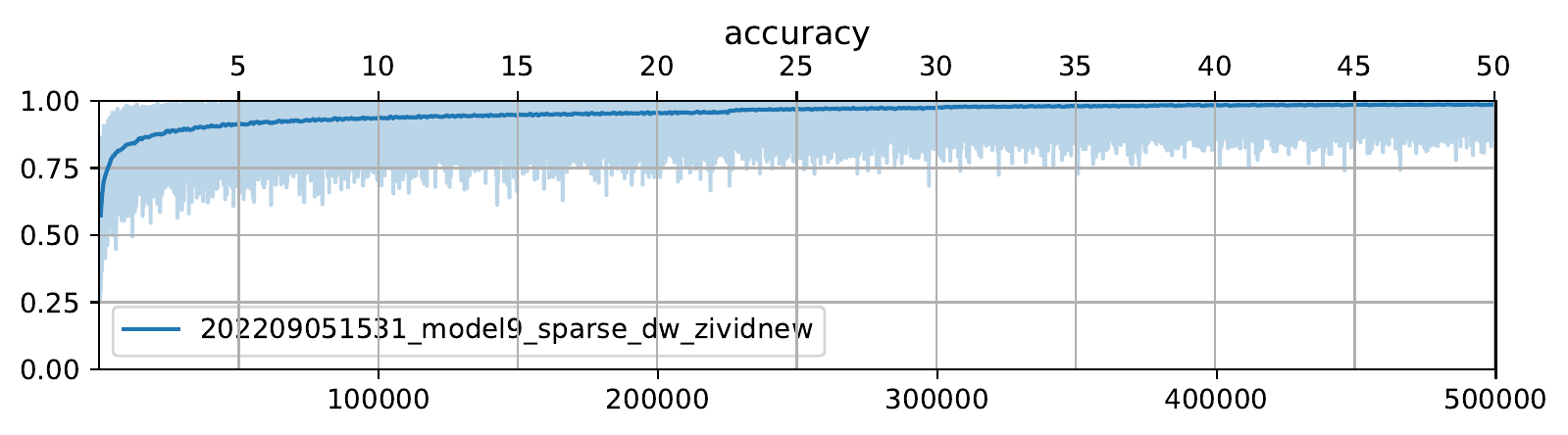}
		\includegraphics[width=\linewidth]{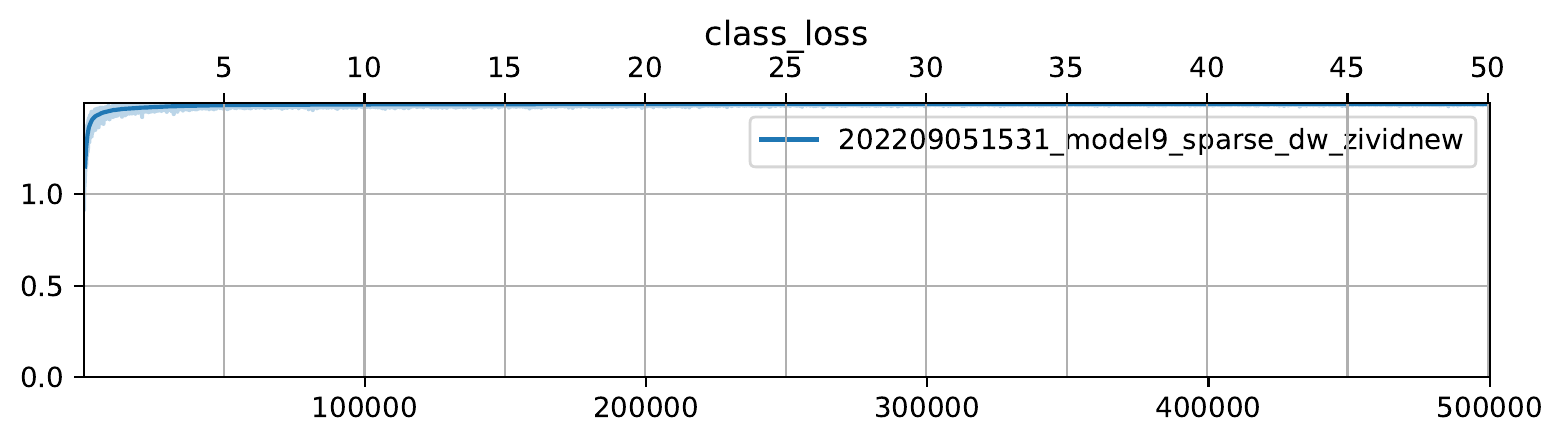}
		\includegraphics[width=\linewidth]{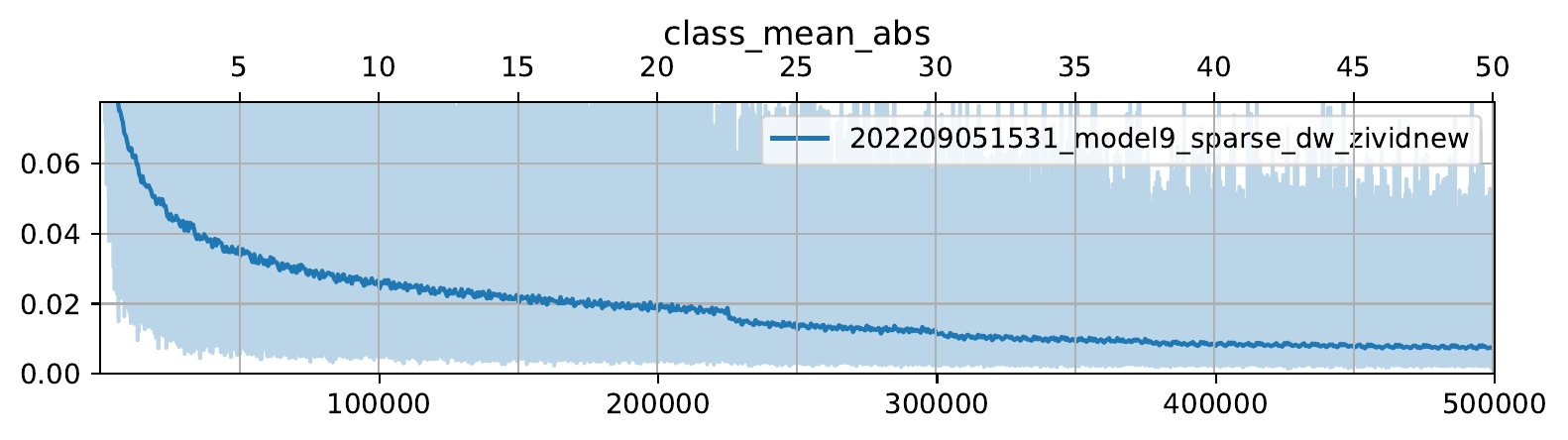}
		\includegraphics[width=\linewidth]{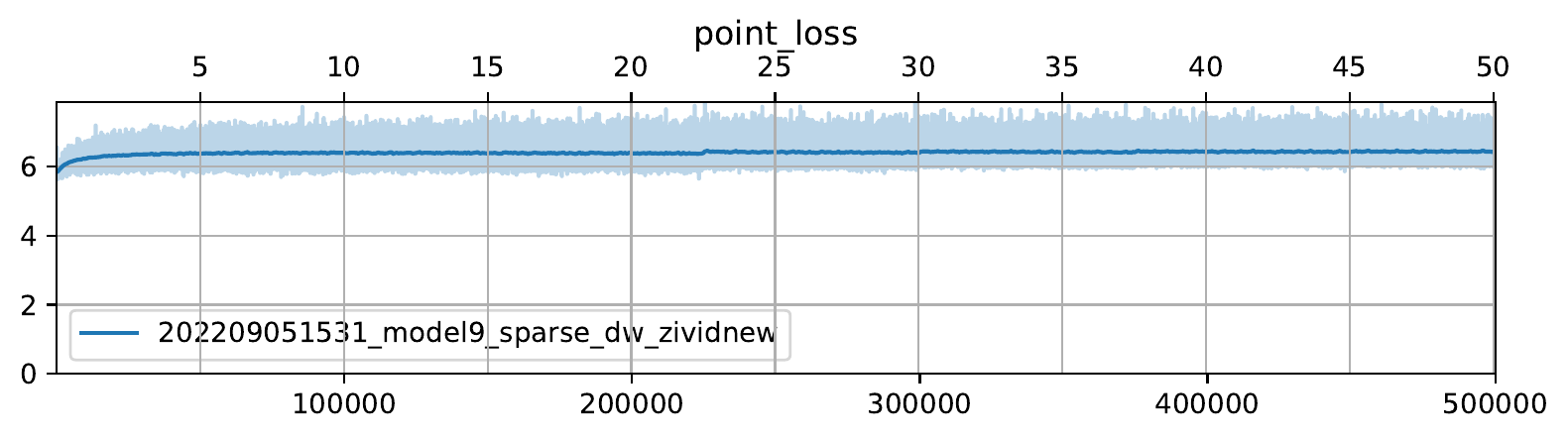}
		\includegraphics[width=\linewidth]{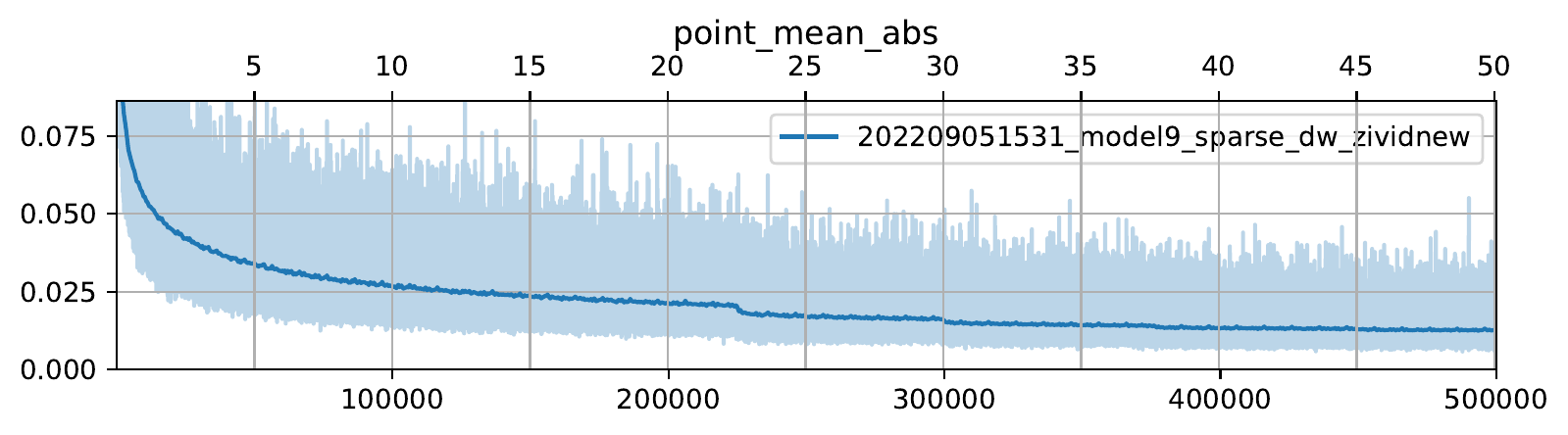}
	\end{subfigure}%
	\begin{subfigure}{.5\linewidth}
		\centering
		\includegraphics[width=\linewidth]{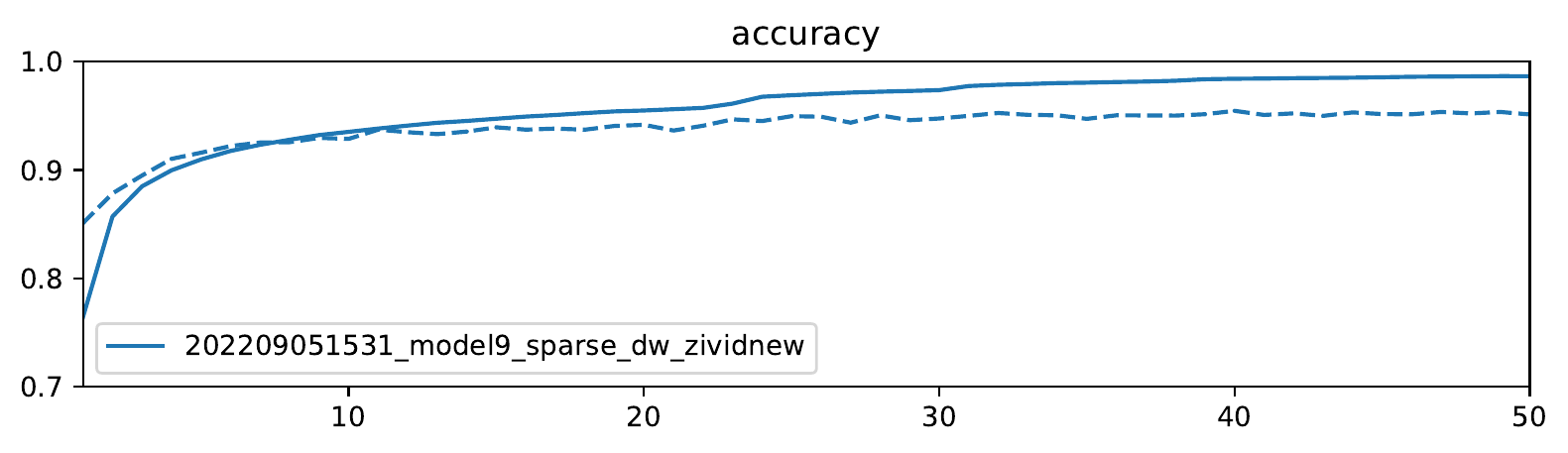}
		\includegraphics[width=\linewidth]{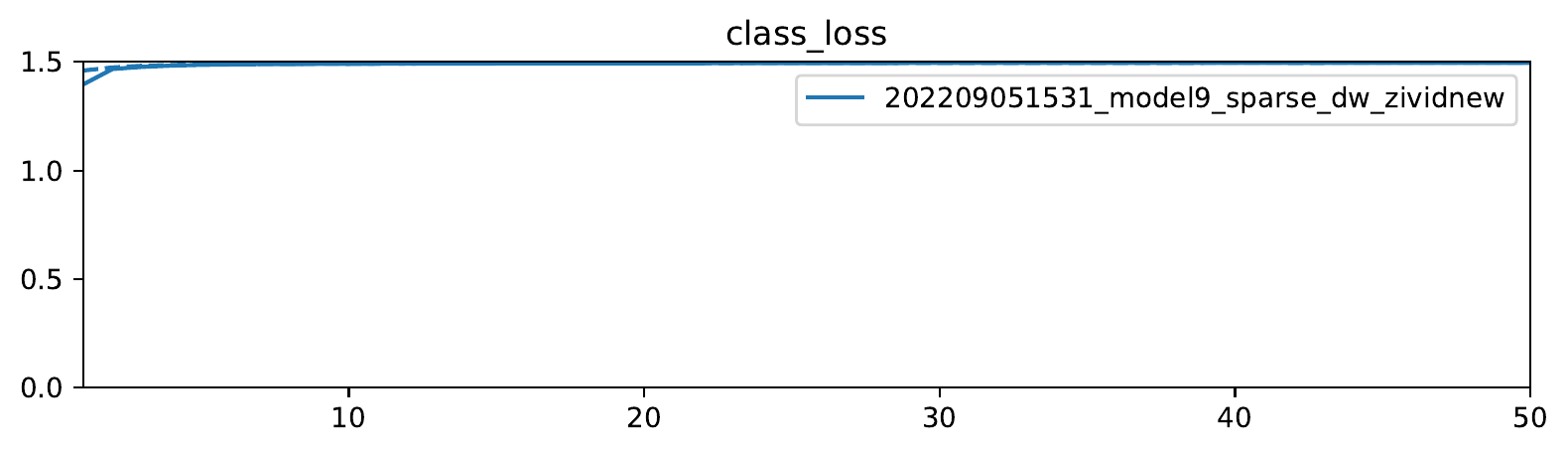}
		\includegraphics[width=\linewidth]{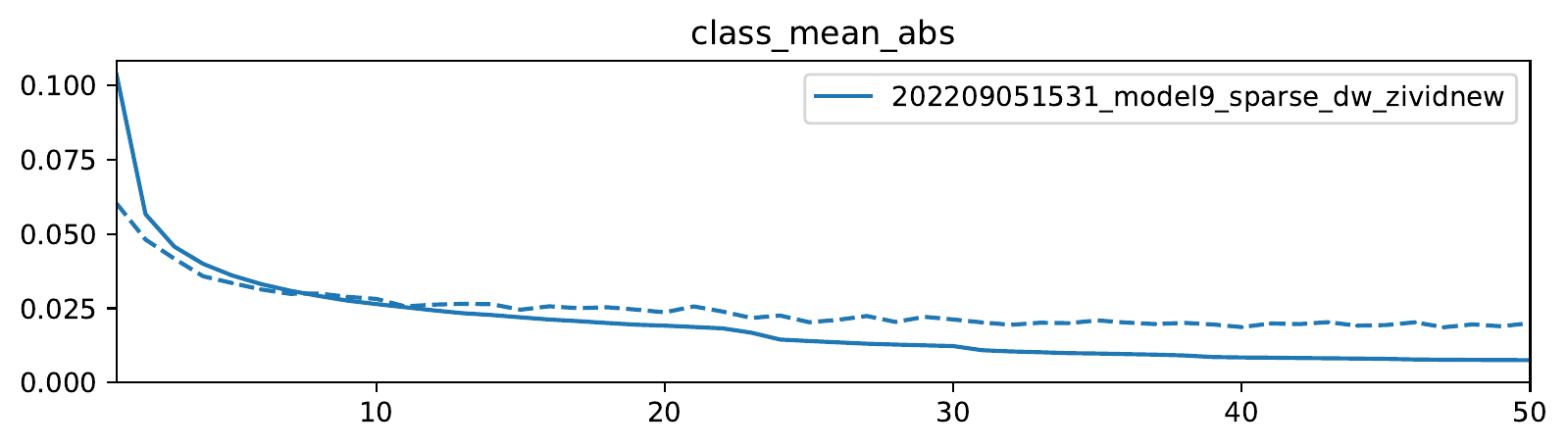}
		\includegraphics[width=\linewidth]{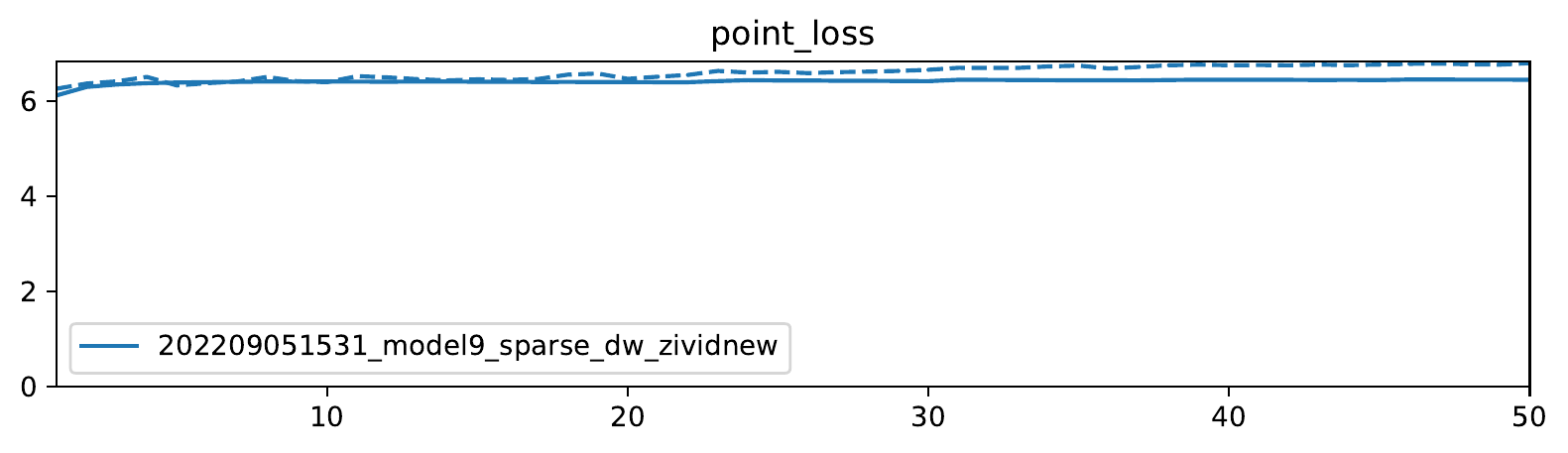}
		\includegraphics[width=\linewidth]{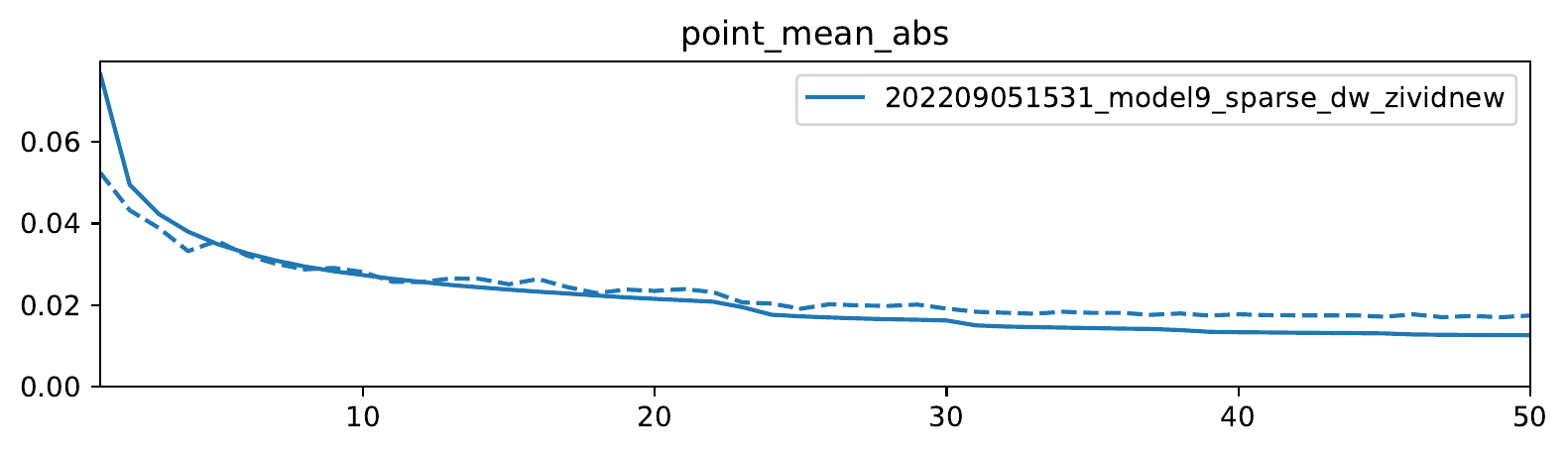}
	\end{subfigure}
	\caption{Training log (left) and history (right); iterations on the bottom, epochs on top; dashed lines represent validation}
	\label{fig:log_history}
\end{figure}

During our work, we trained different sensor specific models and one on orthographic projections which is independent of the intrinsic sensor parameters. Namely, we trained models for the Intel\textregistered~RealSense~D435 which is rather low-cost and for the Zivid~One+~Large which provides sensor data with higher quality.
We trained each model on roughly 120k generated training samples and used 2k for validation.
The training itself was performed on one NVIDIA V100 GPU, with batch size 12 and Adam as optimizer. We started our training with an initial learning rate of $0.001$ and reduced it by half after 225k, 300k, 375k, 450k iterations respectively. Some plots from the training process are shown in \Cref{fig:log_history}.

\iffalse
\begin{figure}[tb]
	\centering
	\includegraphics[width=\linewidth]{images_log_accuracy.png}
	\includegraphics[width=\linewidth]{images_log_class_loss.png}
	\includegraphics[width=\linewidth]{images_log_class_mean_abs.png}
	\includegraphics[width=\linewidth]{images_log_point_loss.png}
	\includegraphics[width=\linewidth]{images_log_point_mean_abs.png}
	\caption{Training log; iterations on the bottom, epochs on top}
	\label{fig:log}
\end{figure}

\begin{figure}[tb]
	\centering
	\includegraphics[width=\linewidth]{images_history_accuracy.pdf}
	\includegraphics[width=\linewidth]{images_history_class_loss.pdf}
	\includegraphics[width=\linewidth]{images_history_class_mean_abs.pdf}
	\includegraphics[width=\linewidth]{images_history_point_loss.pdf}
	\includegraphics[width=\linewidth]{images_history_point_mean_abs.pdf}
	\caption{Training history over epochs; dashed lines are the validation}
	\label{fig:history}
\end{figure}
\fi

\begin{figure}[tb]
	\centering
	\includegraphics[width=\linewidth]{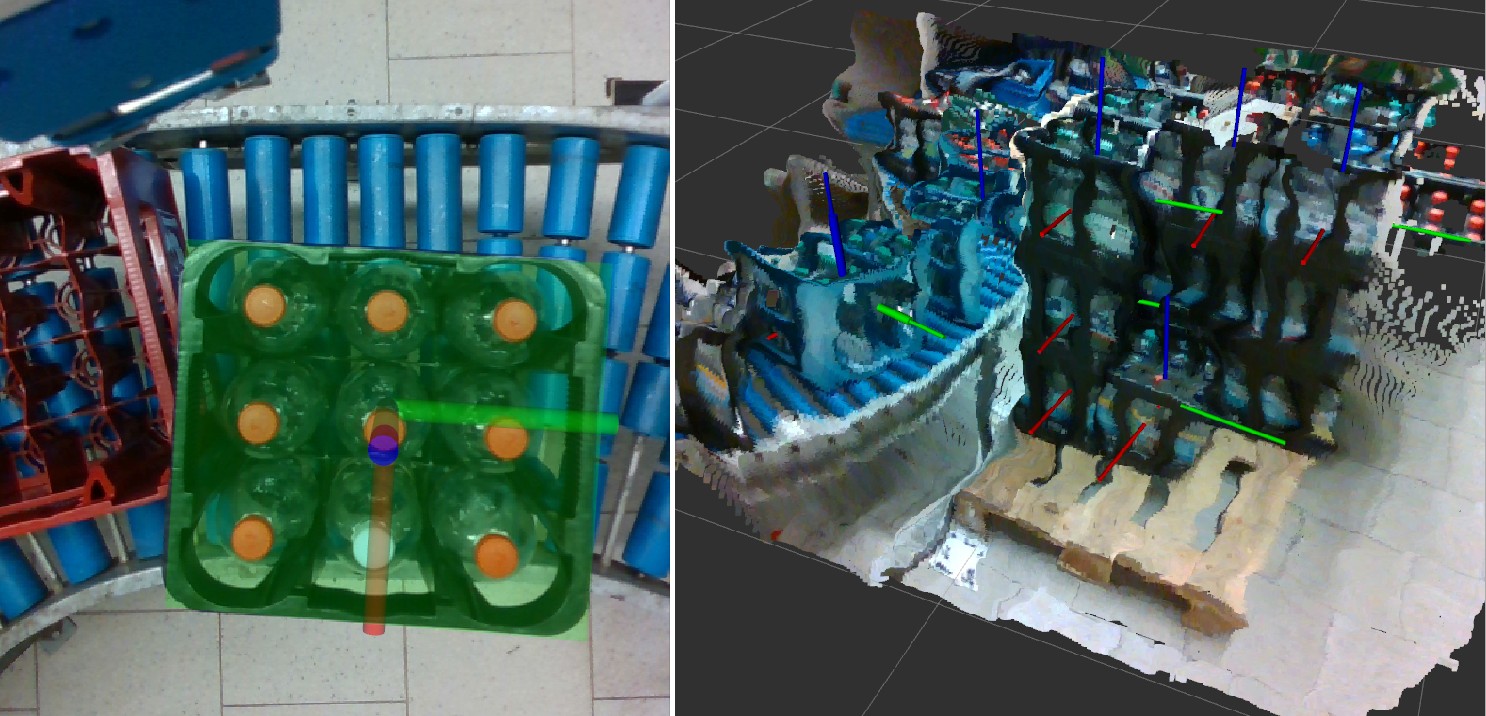}
	\caption{Detection of beverage crates (Intel\textregistered~RealSense~D435)}
	\label{fig:beverage_detection}
\end{figure}

\begin{figure}[tb]
	\begin{subfigure}[b]{\linewidth}
		\includegraphics[width=\linewidth]{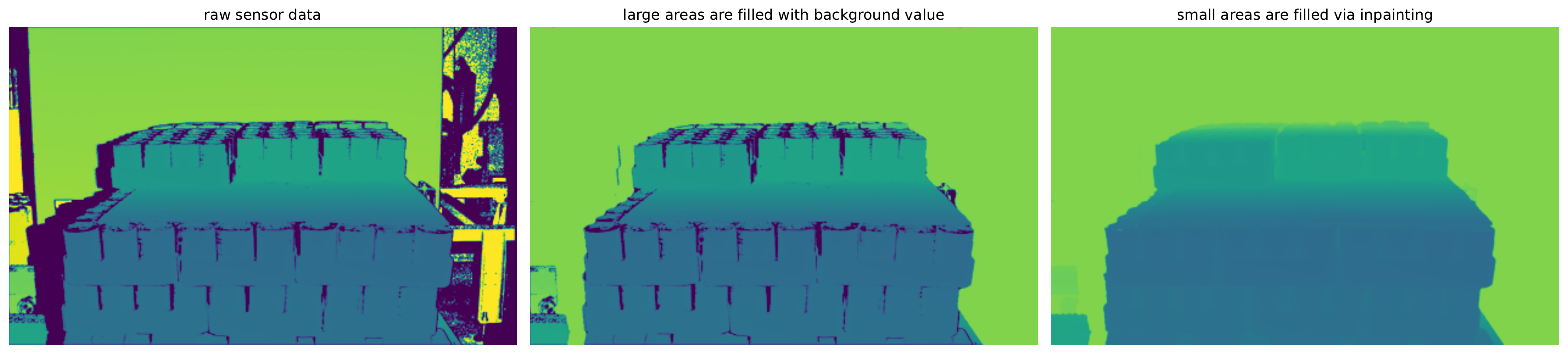}
	\end{subfigure}
	\vfill
	\begin{subfigure}[b]{\linewidth}
		\includegraphics[width=\linewidth]{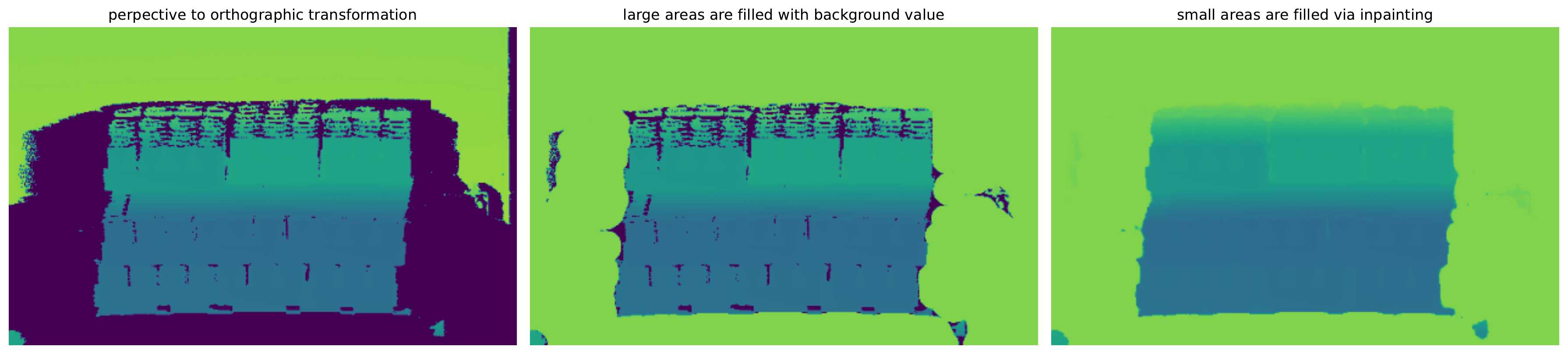}
	\end{subfigure}
	\caption{Our depth completion strategy on raw sensor data (top) and the orthographic projection of the raw sensor data (bottom)}
	\label{fig:inpainting}
\end{figure}

To facilitate the integration and practical use of our detection system, we developed interfaces for ROS and the Universal Robot platform via URCap-XMLRPC and provide a general REST-API. We used the ROS interface to integrate it on a mobile robot platform and applied it to beverage logistics (\Cref{fig:luka,fig:beverage_detection}). We also used the XMLRPC interface to integrate it in our "AI-Picking" lab demonstrator (as shown in the attached video), where we use much smaller pallets and packaging units. There we did not have to train a new model, we simply scaled the whole geometry and worked with an existing model that was trained for euro pallets\footnote{Since there was no small version of the roll-on gripper mentioned above available, we were forced to utilize a suction gripper instead.}.
With the intention to improve the detection performance and for comparison, we implemented as simple strategy for depth map completion specialized for the depalletizing task. Therefore we used morphological operations to set large and far away areas to the depth value of the wall before we use a classical inpainting method following \cite{telea_image_2004} to fill the remaining small areas (\Cref{fig:inpainting}).

% commerial product...

\subsubsection{Evaluation on Real-World Data}

\iffalse
\begin{figure}[tb]
	\centering
	\includegraphics[width=\linewidth]{images_prediction_3d_crop.jpg}
	\caption{Prediction in 3D view; ground truth in green (red), keypointbased regression (blue)}
	\label{fig:prediction_3d}
\end{figure}
\fi

\begin{figure}[tb]
	\centering
	\includegraphics[width=\linewidth]{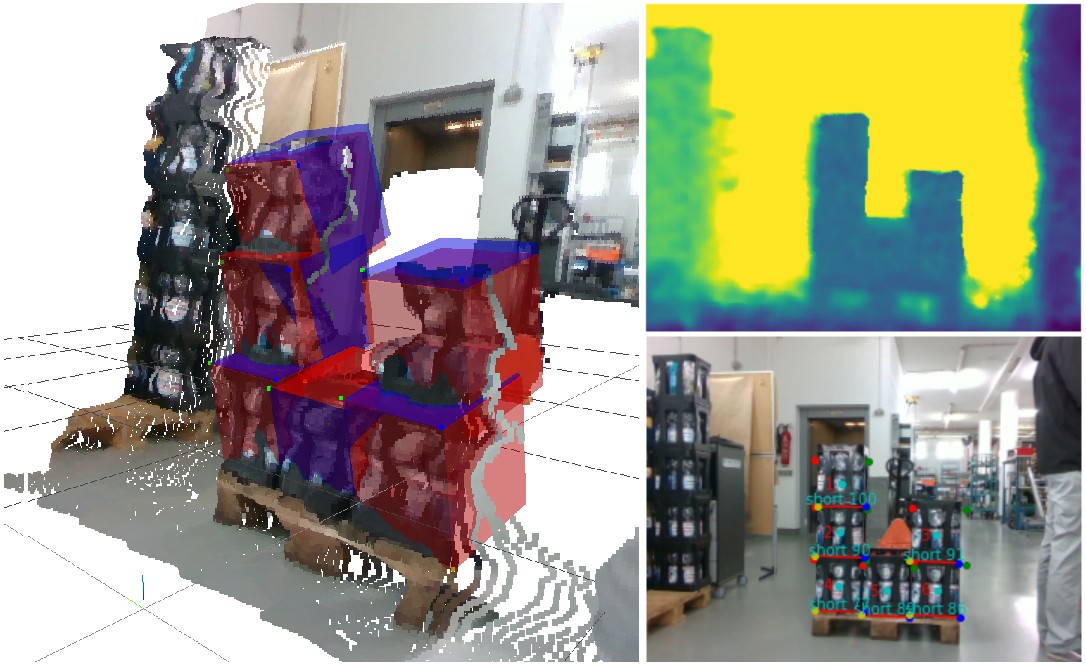}
	\caption{Failure case: bad pose estimations caused by low sensor quality (Intel\textregistered~RealSense~D435); direct regression (red) and based on keypoint regregression (blue)}
	\label{fig:prediction_3d_realsense}
\end{figure}

\begin{figure}[tb]
	\centering
	\includegraphics[width=\linewidth]{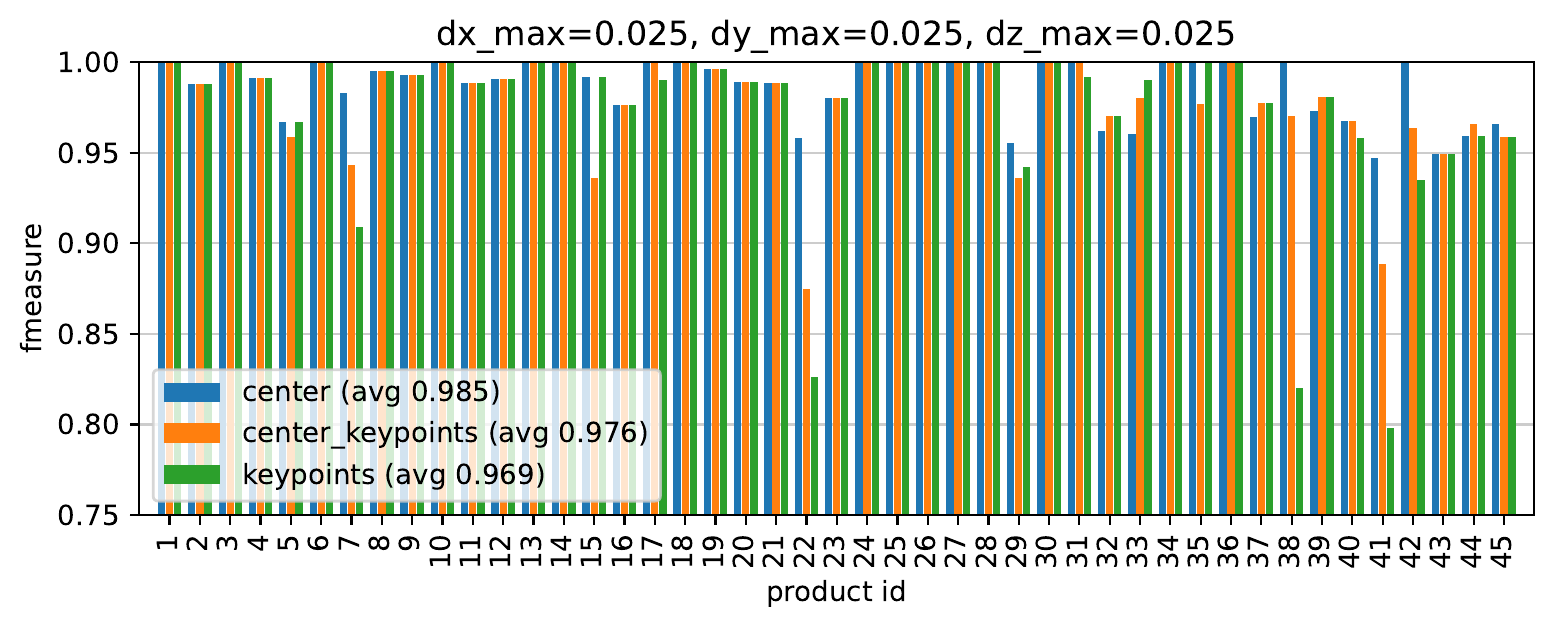}
	\includegraphics[width=\linewidth]{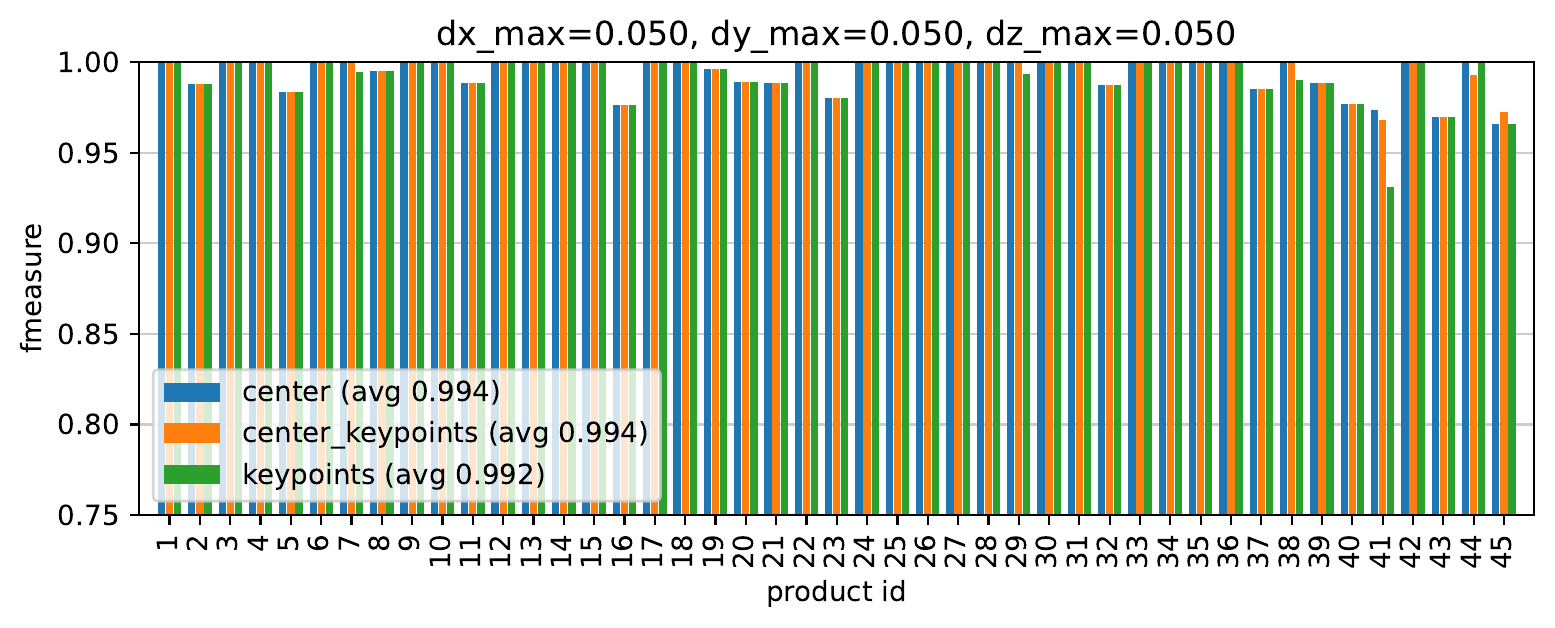}
	\caption{Evaluation on 45 different products with 50 samples per product; maximum permissible deviation 25 mm (top) / 50 mm (bottom)}
	\label{fig:fmeasure}
\end{figure}

\begin{figure}[bt]
	\centering
	\includegraphics[width=\linewidth]{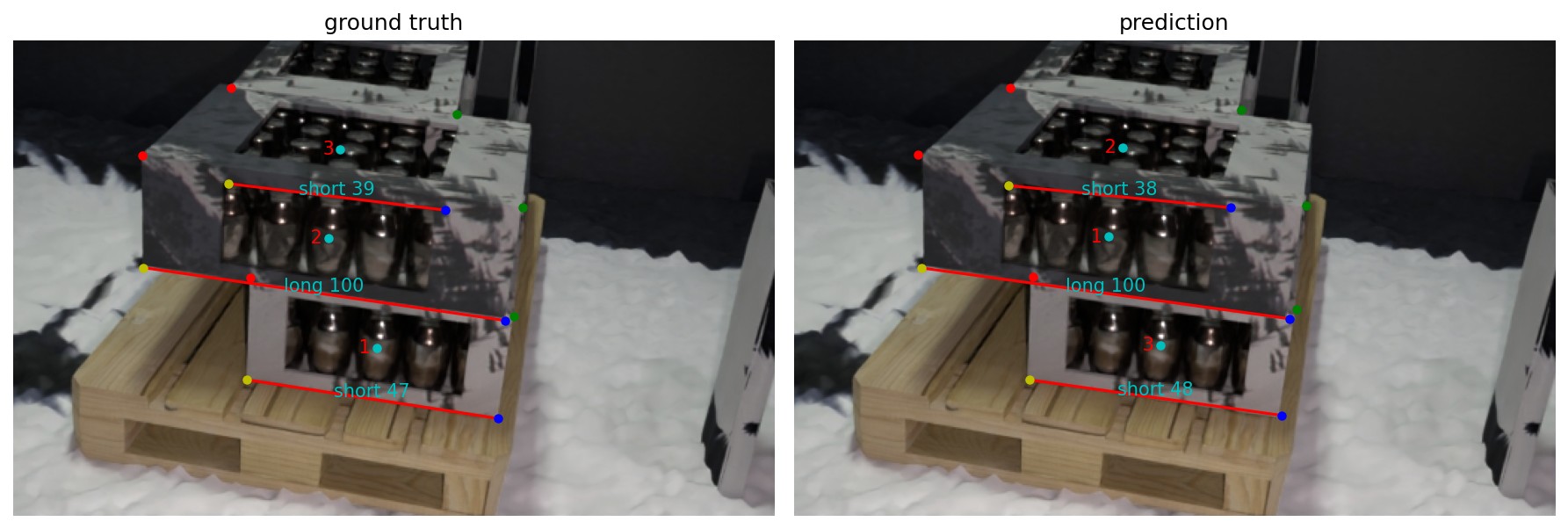}
	\includegraphics[width=\linewidth]{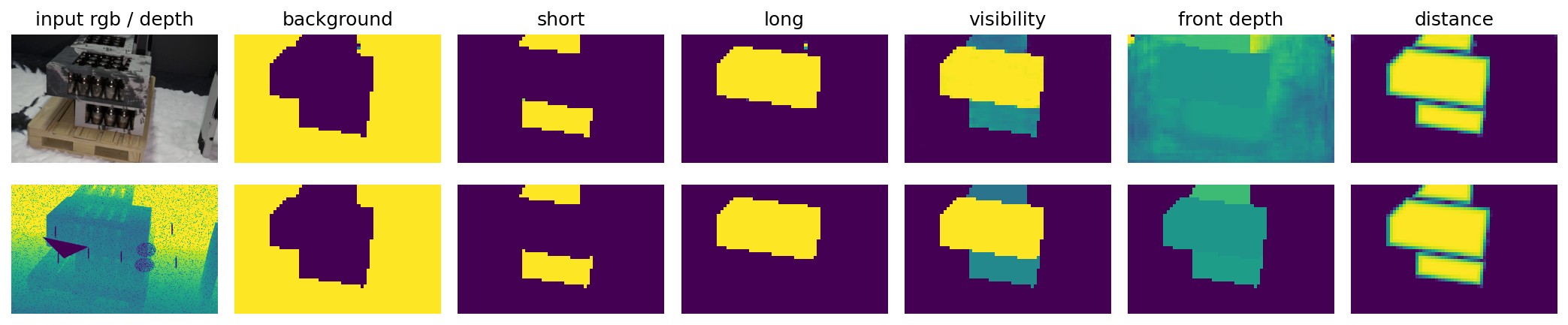}
	\includegraphics[width=\linewidth]{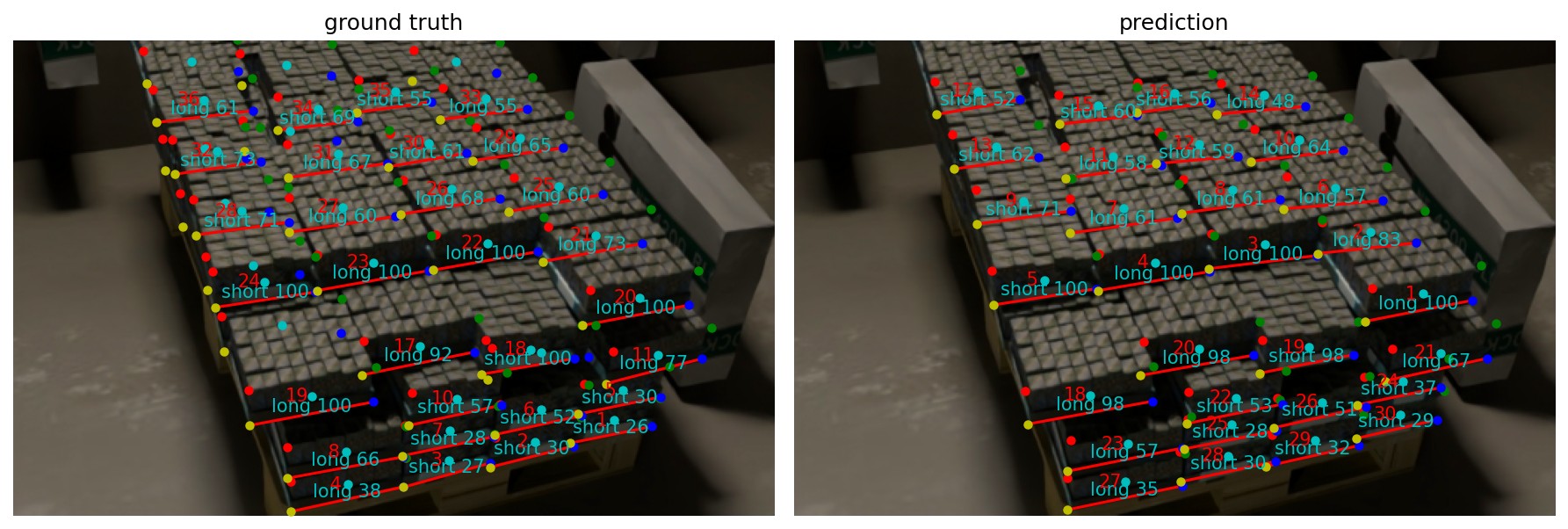}
	\includegraphics[width=\linewidth]{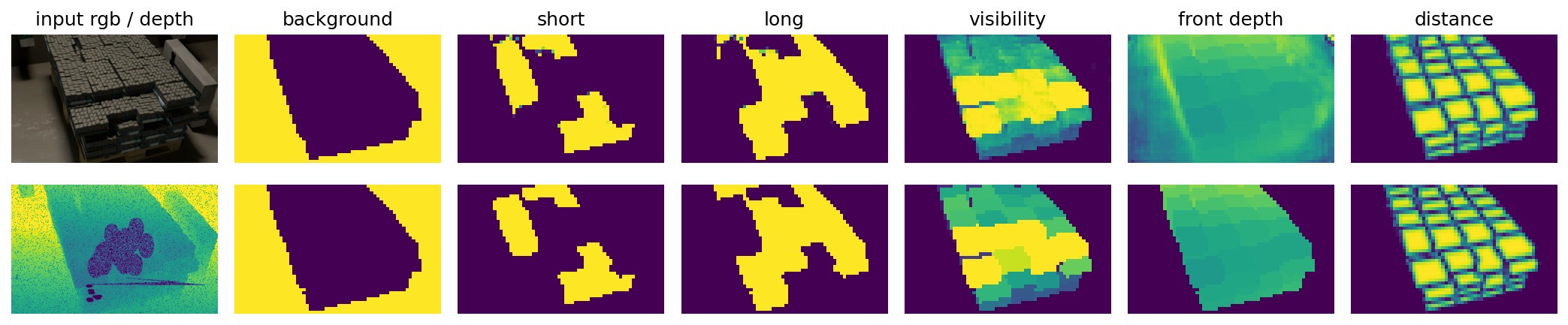}
	\caption{Prediction on generated validation data; only instances with visibility $>0.25$ are shown}
	\label{fig:prediction_sim}
\end{figure}

\begin{figure}[bt]
	\centering
	\includegraphics[width=\linewidth]{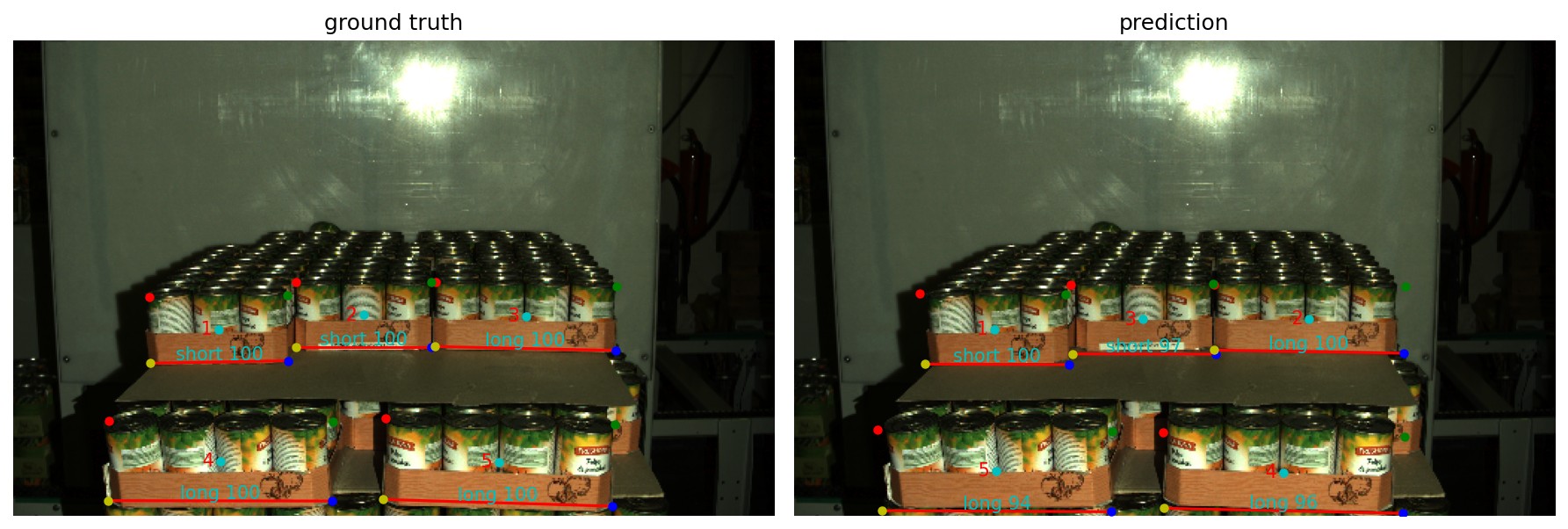}
	\includegraphics[width=\linewidth]{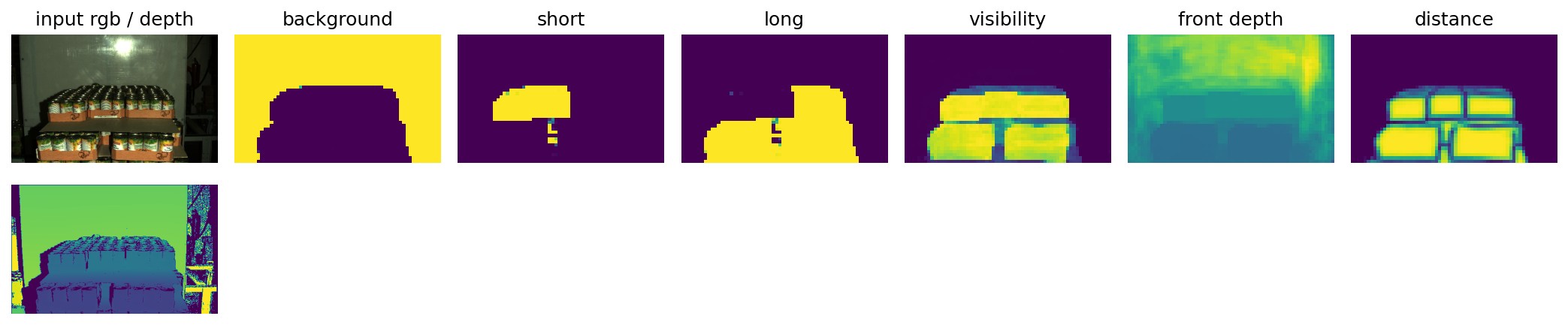}
	\includegraphics[width=\linewidth]{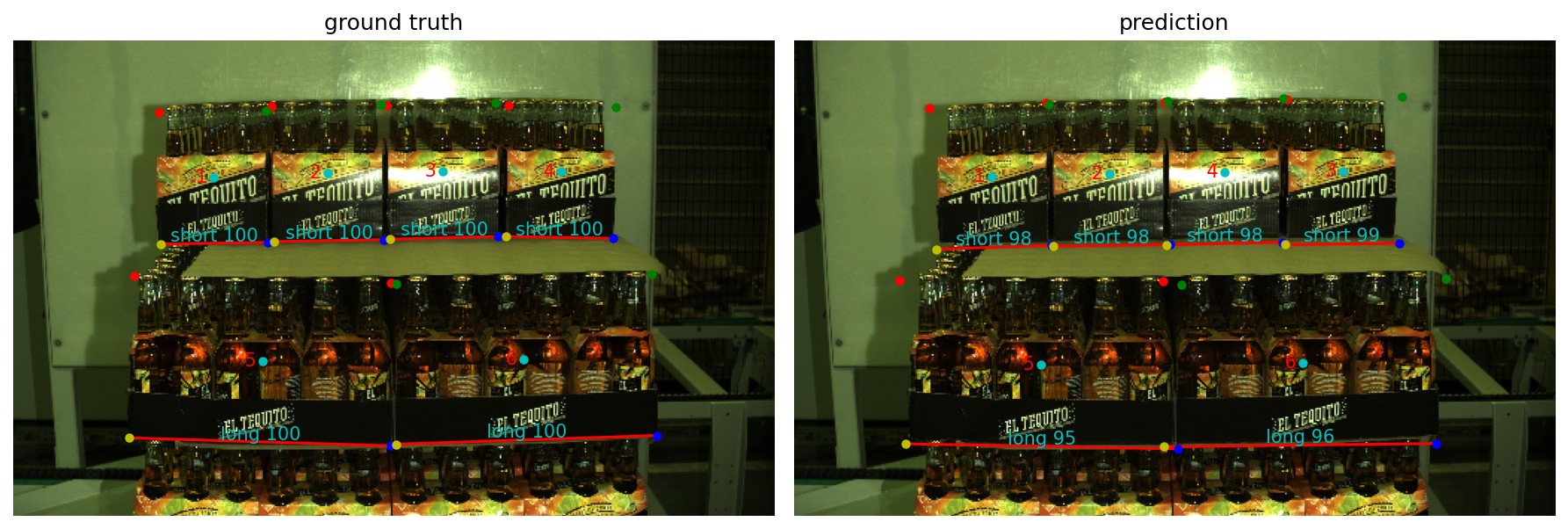}
	\includegraphics[width=\linewidth]{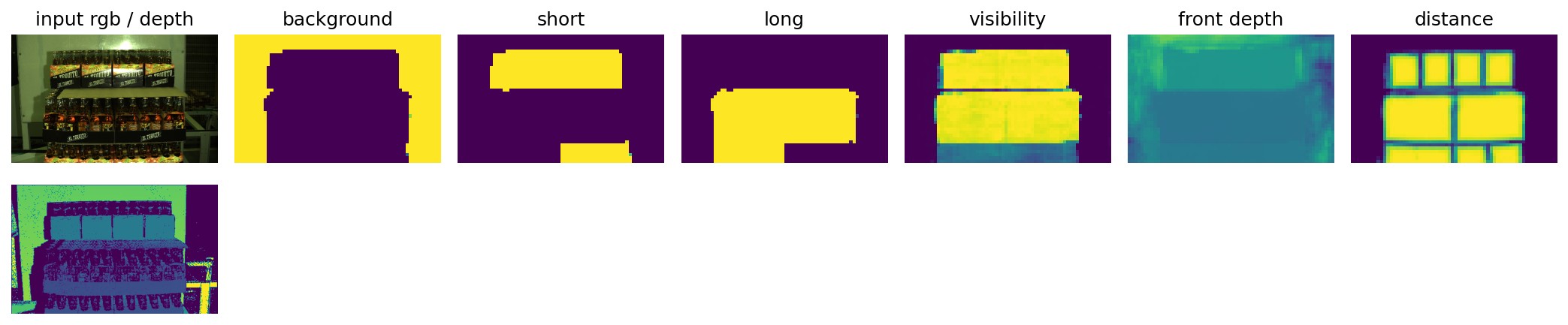}
	\caption{Prediction on real data; only instances with visibility $>0.25$ are shown}
	\label{fig:prediction_real}
	\vspace{-1mm}
\end{figure}

Due to the poor quality of the Intel\textregistered~RealSense~D435, it made little sense for us to annotate larger amounts of data from this sensor that can be used for evaluation.
For the case of the depalettizing task where the box size is known, we got annotations on sensor data from the Zivid~One+~Large.
With this real world annotations we were able evaluate our system on a wide range of different products.
We evaluated it with 50 samples per product on 45 different commercial products (2250 samples in total). Since only the detections in the topmost layer are relevant for the depalletizing task and we only have reliable annotations for this layer, we only detect and evaluate on the topmost layer (8937 instances in total).
We used the \textit{f-measure} (harmonic mean of \textit{precision} and 
\textit{recall}) as primary metric for the evaluation shown in \Cref{fig:fmeasure}. A detection is counted as true positive if its positional distance in pallet frame does not deviate more than a certain maximal value ($d_{x, \text{max}}, d_{y, \text{max}}, d_{z, \text{max}}$), the orientation (short, long) is classified correctly and it is no duplicate detection. We evaluated for the direct regression of the box center, the box center constructed from the 3D keypoints and on the bottom-left and bottom-right front keypoints (red line in \Cref{fig:prediction_sim,fig:prediction_real}, since it is most relevant for picking. With a maximum permissible deviation of 25 mm in $x,y,z$ at an average camera distance of $1.8$~m, we get an average f-measure of 0.985 over all products. The direct regression of the box center location gives the best results for almost all products. We have also done this evaluation with our depth completion strategy from the previous section and found that depth completion roughly decreases the f-measure by 0.01. Closer investigations showed that it fails on large areas without depth information, caused by reflection or transparent materials, which can be better handled by the Partial Convolutions.
The increase of the maximum permissible deviation in position from 25 mm to 50 mm shows that the actual detection of packaging units is not an issue, it is mostly a question of position error.
We found several causes which require more detailed investigation in future work to further reduce the position error:
\begin{enumerate}
	\item Inaccuracy of human annotations in the test data
	\item Low quality of sensor data resulting from transparency and reflections
	\item Small resolution in the model input
	\item Synthetic-to-real gap
	\item Tolerances of provided box size
	\item Deformations of the packaging units
\end{enumerate}

%inaccuracy of human annotations, quality of depth information, prediction error resulting from sim2real

%tried it with and without depth completion, depth completion roughly decreases the fmeasure by 1\%
%sometimes better predictions on left and right edges

%absolute error of the keypoints is average over the batch and spatial dimension but summed over the channel elements

\section{Conclusion and Future Work}
\label{sec:conclusion}
In this paper, we proposed a unified framework for packaging unit detection and applied it to various scenarios. We also contributed with several novel concepts for improving the training of single-shot object detectors and the post-processing stage in \Cref{sec:contributions}.
Resulting from the presented work, directions for further work arise. The proposed \ac{BDT}, \ac{DSSL} and certainty prediction should be applied with several generic state-of-the-art object detection frameworks and evaluated on well known benchmarks like MS COCO \cite{lin_microsoft_2014}.
%It is also not clear at time of writing whether the chosen weighting for the loss terms of the certainty estimate hurt the overall detector performance or not. The proposed network architecture has surely potential to be optimization and so on.

%ablation studies

%\section*{Appendix}

\section*{Acknowledgment}
This work was partially supported by the German Federal Ministry of Education and Research (Deep Picking - Grant No. 01IS20005C) and the Ministry of Economic Affairs, Labour and Tourism of the state Baden-Württemberg (Luka-Beverage - Grant No. 36-3400.7/91, and AI Innovation Center "Learning Systems and Cognitive Robotics").

%\section*{Acronyms}
\begin{acronym}
    \setlength{\parskip}{0ex}
    \setlength{\itemsep}{0.4ex}
    \acro{AD}{Automatic Differentiation}
    \acro{AI}{Artificial Intelligence}
    \acro{ANN}{Artificial Neural Network}
    \acro{AP}{Average Precision}
    \acro{API}{Application Programming Interface}
    \acro{AR}{Augmented Reality}
    \acro{ASIC}{Application-Specific Integrated Circuit}
    \acro{BCE}{Binary Cross Entropy}
    \acro{BLSTM}{Bidirectional Long Short-Term Memory}
    \acro{BN}{Batch Normalization}
    \acro{BPTT}{Backpropagation Through Time}
    \acro{BRNN}{Bidirectional Recurrent Neural Networks}
    \acro{CE}{Cross Entropy}
    \acro{CNN}{Convolutional Neural Network}
    \acro{CPU}{Central Processing Unit}
    \acro{CRNN}{Convolutional Recurrent Neural Network}
    \acro{CTC}{Connectionist Temporal Classification}
    \acro{CV}{Computer Vision}
    \acro{DCN}{Deep Convolutional Network}
    \acro{DCNN}{Deep Convolutional Neural Network}
    \acro{DL}{Deep Learning}
    \acro{DNA}{Deoxyribonucleic acid}
    \acro{DNN}{Deep Neural Network}
    \acro{DPM}{Deformable Part Models}
    \acro{CNTK}{Computational Network Toolkit}
    \acro{FC}{Fully Connected}
    \acro{FCN}{Fully Convolutional Network}
    \acro{FFNN}{Feed-Forward Neural Network}
    \acro{FL}{Focal Loss}
    \acro{FN}{False Negative}
    \acro{FP}{False Positive}
    \acro{FPGA}{Field-Programmable Gate Array}
    \acro{GAN}{Generative Adversarial Network}
    \acro{GD}{Gradient Descent}
    \acro{GIMP}{GNU Image Manipulation Program}
    \acro{GPU}{Graphics Processing Unit}
    \acro{GRU}{Gated Recurrent Unit}
    \acro{HMM}{Hidden Markov Model}
    \acro{HED}{Holistically-Nested Edge Detection}
    \acro{HOG}{Histogram of Oriented Gradients}
    \acro{IoU}{Intersection over Union}
    \acro{IoT}{Internet of Things}
    \acro{JVM}{Java Virtual Machine}
    \acro{JIT}{just-in-time}
    \acro{KL}{Kullback Leibler}
    \acro{LSTM}{Long Short-Term Memory}
    \acro{mAP}{mean Average Precision}
    \acro{MLP}{Multilayer Perceptron}
    \acro{MAE}{Mean Average Error}
    \acro{MSER}{Maximally Stable Extremal Regions}
    \acro{MLE}{Maximum Likelihood Estimation}
    \acro{NLP}{Natural Language Processing}
    \acro{MSE}{Mean Square Error}
    \acro{NMS}{Non Maximum Suppression}
    \acro{NN}{Neural Network}
    \acro{OCR}{Optical Character Recognition}
    \acro{OHEM}{Online Hard Example Mining}
    \acro{OHNM}{Online Hard Negative Mining}
    \acro{ONNX}{Open Neural Network Exchange}
    \acro{OS}{Operating System}
    \acro{PBR}{Physically Based Rendering}
    \acro{PReLU}{Parametric Rectified Linear Unit}
    \acro{RBF}{Radial Basis Function}
    \acro{R-CNN}{Region-based Convolutional Neural Netwoks}
    \acro{RFL}{Reduced Focal Loss}
    \acro{ReLU}{Rectified Linear Unit}
    \acro{R-FCN}{Region-based Fully Convolutional Network}
    \acro{RGB}{Red, Green, Blue}
    \acro{RNN}{Recurrent Neural Network}
    \acro{RoI}{Region of Interest}
    \acro{RPN}{Region Proposal Network}
    \acro{ROS}{Robot Operating System}
    \acro{SGD}{Stochastic Gradient Descent}
    \acro{SL}{Shrinkage Loss}
    \acro{SSD}{Single Shot MultiBox Detector}
    \acro{TPS}{Thin-Plate Spline}
    \acro{STN}{Spatial Transformer Network}
    \acro{SVM}{Support Vector Machine}
    \acro{TN}{True Negative}
    \acro{TP}{True Positive}
    \acro{TPU}{Tensor Processing Unit}
    \acro{YOLO}{You Only Look Once}
    
    \acro{BDT}{Bounded Distance Transform}
    \acro{DSSL}{Dynamically Scaled Shrinkage Loss}
    
    %DSOD CRNN
\end{acronym}

\bibliographystyle{IEEEtran}

%\bibliography{IEEEabrv,references}
\bibliography{references}

\begin{thebibliography}{10}
\providecommand{\url}[1]{#1}
\csname url@rmstyle\endcsname
\providecommand{\newblock}{\relax}
\providecommand{\bibinfo}[2]{#2}
\providecommand\BIBentrySTDinterwordspacing{\spaceskip=0pt\relax}
\providecommand\BIBentryALTinterwordstretchfactor{4}
\providecommand\BIBentryALTinterwordspacing{\spaceskip=\fontdimen2\font plus
\BIBentryALTinterwordstretchfactor\fontdimen3\font minus
  \fontdimen4\font\relax}
\providecommand\BIBforeignlanguage[2]{{%
\expandafter\ifx\csname l@#1\endcsname\relax
\typeout{** WARNING: IEEEtran.bst: No hyphenation pattern has been}%
\typeout{** loaded for the language `#1'. Using the pattern for}%
\typeout{** the default language instead.}%
\else
\language=\csname l@#1\endcsname
\fi
#2}}

\bibitem{muetherich_gripping_2010}
H.~Muetherich, F.~Simons, and A.~Verl, ``Gripping systems for intralogistics -
  aiming at the "swiss army knife" of intralogistics solutions,'' in
  \emph{{ISR} and {ROBOTIK}}, 2010, pp. 1--7.

\bibitem{hoque_comprehensive_2021}
S.~Hoque, M.~Y. Arafat, S.~Xu, A.~Maiti, and Y.~Wei, ``A comprehensive review
  on 3d object detection and 6d pose estimation with deep learning,''
  \emph{{IEEE} Access}, vol.~9, pp. 143\,746--143\,770, 2021.

\bibitem{gorschluter_survey_2022}
F.~Gorschlüter, P.~Rojtberg, and T.~Pöllabauer, ``A survey of 6d object
  detection based on 3d models for industrial applications,'' \emph{Journal of
  Imaging}, vol.~8, no.~3, p.~53, 2022.

\bibitem{monica_detection_2020}
R.~Monica, J.~Aleotti, and D.~L. Rizzini, ``Detection of parcel boxes for
  pallet unloading using a 3d time-of-flight industrial sensor,'' in
  \emph{{IEEE} Internat. Conf. on Robotic Computing ({IRC})}, 2020, pp.
  314--318.

\bibitem{aleotti_toward_2021}
J.~Aleotti, A.~Baldassarri, M.~Bonfè, M.~Carricato, D.~Chiaravalli, and {et
  al.}, ``Toward future automatic warehouses: An autonomous depalletizing
  system based on mobile manipulation and 3d perception,'' \emph{Applied
  Sciences}, vol.~11, no.~13, p. 5959, 2021.

\bibitem{holz_real_time_2015}
D.~Holz, A.~Topalidou-Kyniazopoulou, J.~Stückler, and S.~Behnke, ``Real-time
  object detection, localization and verification for fast robotic
  depalletizing,'' in \emph{{IROS}}, 2015, pp. 1459--1466.

\bibitem{arpenti_rgbd_2020}
P.~Arpenti, R.~Caccavale, G.~Paduano, G.~Andrea~Fontanelli, V.~Lippiello,
  L.~Villani, and B.~Siciliano, ``{RGB}-d recognition and localization of cases
  for robotic depalletizing in supermarkets,'' \emph{{IEEE} Robotics and
  Automation Letters}, vol.~5, no.~4, pp. 6233--6238, 2020.

\bibitem{li_workpiece_2020}
J.~Li, J.~Kang, Z.~Chen, F.~Cui, and Z.~Fan, ``A workpiece localization method
  for robotic de-palletizing based on region growing and {PPHT},'' \emph{{IEEE}
  Access}, vol.~8, pp. 166\,365--166\,376, 2020.

\bibitem{prasse_concept_2011}
C.~Prasse, S.~Skibinski, F.~Weichert, J.~Stenzel, H.~Müller, and M.~ten
  Hompel, ``Concept of automated load detection for de-palletizing using depth
  images and {RFID} data,'' in \emph{{ICCSCE}}, 2011, pp. 249--254.

\bibitem{caccavale_flexible_2020}
R.~Caccavale, P.~Arpenti, G.~Paduano, A.~Fontanellli, V.~Lippiello, L.~Villani,
  and B.~Siciliano, ``A flexible robotic depalletizing system for supermarket
  logistics,'' \emph{{IEEE} Robotics and Automation Letters}, vol.~5, no.~3,
  pp. 4471--4476, 2020.

\bibitem{liu_ssd_2016}
W.~Liu, D.~Anguelov, D.~Erhan, C.~Szegedy, S.~Reed, C.-Y. Fu, and A.~C. Berg,
  ``{SSD}: Single shot {MultiBox} detector,'' \emph{{arXiv}:1512.02325 [cs]},
  vol. 9905, pp. 21--37, 2016.

\bibitem{bochkovskiy_yolov4_2020}
A.~Bochkovskiy, C.-Y. Wang, and H.-Y.~M. Liao, ``{YOLOv}4: Optimal speed and
  accuracy of object detection,'' \emph{{arXiv}:2004.10934 [cs]}, 2020.

\bibitem{wang_yolov7_2022}
C.-Y. Wang, A.~Bochkovskiy, and H.-Y.~M. Liao, ``{YOLOv}7: Trainable
  bag-of-freebies sets new state-of-the-art for real-time object detectors,''
  2022.

\bibitem{lin_focal_2017}
T.-Y. Lin, P.~Goyal, R.~Girshick, K.~He, and P.~Dollár, ``Focal loss for dense
  object detection,'' \emph{{arXiv}:1708.02002 [cs]}, 2017.

\bibitem{shrivastava_training_2016}
A.~Shrivastava, A.~Gupta, and R.~Girshick, ``Training region-based object
  detectors with online hard example mining,'' \emph{{arXiv}:1604.03540 [cs]},
  2016.

\bibitem{sergievskiy_reduced_2019}
N.~Sergievskiy and A.~Ponamarev, ``Reduced focal loss: 1st place solution to
  {xView} object detection in satellite imagery,'' \emph{{arXiv}:1903.01347
  [cs]}, 2019.

\bibitem{lu_deep_2018}
X.~Lu, C.~Ma, B.~Ni, X.~Yang, I.~Reid, and M.-H. Yang, ``Deep regression
  tracking with shrinkage loss,'' p.~17, 2018.

\bibitem{huber_robust_1964}
P.~J. Huber, ``Robust estimation of a location parameter,'' \emph{The Annals of
  Mathematical Statistics}, vol.~35, no.~1, pp. 73--101, 1964.

\bibitem{girshick_fast_2015}
R.~Girshick, ``Fast r-{CNN},'' \emph{{arXiv}:1504.08083 [cs]}, 2015.

\bibitem{zheng_distance-iou_2019}
Z.~Zheng, P.~Wang, W.~Liu, J.~Li, R.~Ye, and D.~Ren, ``Distance-{IoU} loss:
  Faster and better learning for bounding box regression,''
  \emph{{arXiv}:1911.08287 [cs]}, 2019.

\bibitem{tian_fcos_2019}
Z.~Tian, C.~Shen, H.~Chen, and T.~He, ``{FCOS}: Fully convolutional one-stage
  object detection,'' 2019.

\bibitem{wu_iou-aware_2020}
S.~Wu, X.~Li, and X.~Wang, ``{IoU}-aware single-stage object detector for
  accurate localization,'' 2020.

\bibitem{chollet_xception_2016}
F.~Chollet, ``Xception: Deep learning with depthwise separable convolutions,''
  \emph{{arXiv}:1610.02357 [cs]}, 2016.

\bibitem{howard_mobilenets_2017}
A.~G. Howard, M.~Zhu, B.~Chen, D.~Kalenichenko, W.~Wang, T.~Weyand,
  M.~Andreetto, and H.~Adam, ``{MobileNets}: Efficient convolutional neural
  networks for mobile vision applications,'' \emph{{arXiv}:1704.04861 [cs]},
  2017.

\bibitem{sandler_mobilenetv2_2018}
M.~Sandler, A.~Howard, M.~Zhu, A.~Zhmoginov, and L.-C. Chen, ``{MobileNetV}2:
  Inverted residuals and linear bottlenecks,'' 2018.

\bibitem{liu_intriguing_2018}
R.~Liu, J.~Lehman, P.~Molino, F.~P. Such, E.~Frank, A.~Sergeev, and
  J.~Yosinski, ``An intriguing failing of convolutional neural networks and the
  {CoordConv} solution,'' \emph{{arXiv}:1807.03247 [cs, stat]}, 2018.

\bibitem{uhrig_sparsity_2017}
J.~Uhrig, N.~Schneider, L.~Schneider, U.~Franke, T.~Brox, and A.~Geiger,
  ``Sparsity invariant {CNNs},'' \emph{{arXiv}:1708.06500 [cs]}, 2017.

\bibitem{liu_image_2018}
G.~Liu, F.~A. Reda, K.~J. Shih, T.-C. Wang, A.~Tao, and B.~Catanzaro, ``Image
  inpainting for irregular holes using partial convolutions,'' 2018.

\bibitem{hinterstoisser_annotation_2019}
S.~Hinterstoisser, O.~Pauly, H.~Heibel, M.~Marek, and M.~Bokeloh, ``An
  annotation saved is an annotation earned: Using fully synthetic training for
  object instance detection,'' 2019.

\bibitem{hodan_photorealistic_2019}
T.~Hodan, V.~Vineet, R.~Gal, E.~Shalev, J.~Hanzelka, T.~Connell, P.~Urbina,
  S.~N. Sinha, and B.~Guenter, ``Photorealistic image synthesis for object
  instance detection,'' 2019.

\bibitem{devlin_bert_2019}
J.~Devlin, M.-W. Chang, K.~Lee, and K.~Toutanova, ``{BERT}: Pre-training of
  deep bidirectional transformers for language understanding,''
  \emph{{arXiv}:1810.04805 [cs]}, 2019.

\bibitem{he_masked_2021}
K.~He, X.~Chen, S.~Xie, Y.~Li, P.~Dollár, and R.~Girshick, ``Masked
  autoencoders are scalable vision learners,'' 2021.

\bibitem{srivastava_dropout_2014}
N.~Srivastava, G.~E. Hinton, A.~Krizhevsky, I.~Sutskever, and R.~Salakhutdinov,
  ``Dropout: a simple way to prevent neural networks from overfitting.''
  \emph{Journal of machine learning research}, vol.~15, no.~1, pp. 1929--1958,
  2014.

\bibitem{strutz_distance_2021}
T.~Strutz, ``The distance transform and its computation,'' 2021.

\bibitem{noauthor_opencv_2014}
\emph{The {OpenCV} Reference Manual}, 2nd~ed.\hskip 1em plus 0.5em minus
  0.4em\relax Itseez, 2014.

\bibitem{telea_image_2004}
A.~Telea, ``An image inpainting technique based on the fast marching method,''
  \emph{Journal of Graphics Tools}, vol.~9, no.~1, pp. 23--34, 2004.

\bibitem{lin_microsoft_2014}
T.-Y. Lin, M.~Maire, S.~Belongie, L.~Bourdev, R.~Girshick, J.~Hays, P.~Perona,
  D.~Ramanan, C.~L. Zitnick, and P.~Dollár, ``Microsoft {COCO}: Common objects
  in context,'' \emph{{arXiv}:1405.0312 [cs]}, 2014.

\end{thebibliography}

\end{document}